\documentclass{article}
\usepackage{iclr2027_conference,times}
\usepackage{amsmath,amsfonts,bm}

\def\eqref#1{equation~\ref{#1}}
\def\1{\bm{1}}

\DeclareMathAlphabet{\mathsfit}{\encodingdefault}{\sfdefault}{m}{sl}
\SetMathAlphabet{\mathsfit}{bold}{\encodingdefault}{\sfdefault}{bx}{n}

\usepackage[utf8]{inputenc}
\usepackage[T1]{fontenc}
\usepackage{microtype}
\usepackage{booktabs}
\usepackage{array}
\usepackage{graphicx}
\usepackage{caption}
\usepackage{float}
\usepackage{enumitem}
\usepackage{hyperref}
\usepackage{url}
\usepackage{seqsplit}

\newsavebox{\fitbox}
\newcommand{\fitwidth}[1]{%
  \sbox{\fitbox}{#1}%
  \ifdim\wd\fitbox>\linewidth
    \resizebox{\linewidth}{!}{\usebox{\fitbox}}%
  \else
    \usebox{\fitbox}%
  \fi}

\title{Decision-FLIP: Budget-Free Structured Pruning for Gaussian Occupancy Models}

\author{
  \makebox[\textwidth][c]{\textbf{Jingxing Li}$^*$ \quad \textbf{Anurag Paul} \quad \textbf{Subhasis Das} \quad \textbf{Cindy Xu}$^\dagger$} \\[0.4em]
  \makebox[\textwidth][c]{Zoox Inc.} \\[0.4em]
  \makebox[\textwidth][c]{\texttt{\{jinli,apaul,subhasis,cxu\}@zoox.com}}
}
\iclrfinalcopy
\begin{document}
\maketitle
\lhead{}

\begin{abstract}
Occupancy predictors rely on large persistent banks of Gaussians or learned queries whose size is fixed by hand. Common pruning proxies, such as opacity, activation frequency, or parameter magnitude, do not measure whether removing a row changes the model's voxel decisions. We introduce \textbf{Decision-FLIP}, a budget-free compilation framework whose scoring and structure selection need only unlabeled calibration inputs. Exact Mean-FLIP scores each persistent row by the number of voxel argmax decisions that change when the row is removed from the readout; a guarded Otsu split of the log-score distribution selects the survivor count without a target and abstains when the split is unsupported; and the survivors are physically compiled into a smaller checkpoint, so camera sampling, attention, and aggregation shrink with the bank. For streaming banks, scores are integrated over each row's lifecycle through warp, death, and refill. Across four Gaussian-bank model-dataset settings and one persistent-query transfer, the automatically selected cuts remove \textbf{12--79\%} of eligible rows and cut bank/query-dependent latency by \textbf{7--45\%} across settings with component profiles, with whole-model savings of up to \textbf{8.0\% on the primary endpoints} bounded by fixed-cost image, voxel, and sensor modules and peak-memory savings of up to \textbf{16.6\%}. Under matched one-epoch recovery controls, semantic mIoU changes by at most 0.18 point. Same-$K$ controls show that the ranking, not only the bank size, matters: random removal collapses GaussianFormer-1 and OccFormer, while label-free Mean-FLIP matches or outperforms supervised Taylor ranking on all five settings where that control was run. The method applies when removable primitives keep a stable checkpoint-row identity; it is budget-free in survivor count, not parameter-free.
\end{abstract}

\section{Introduction}

Real-time 3D occupancy prediction for autonomous driving must balance spatial coverage against latency and memory. Sparse Gaussian occupancy models replace dense 3D feature volumes with a bank of learned scene primitives. The attributes of each Gaussian depend on the input, but the number of generator slots is usually an architecture constant chosen by hand. Larger banks improve coverage at the cost of encoder, camera-sampling, and aggregation time. Activity criteria such as opacity or fire rate do not answer the question that matters for structural pruning: would removing this primitive change the final voxel decision? A highly opaque or frequently active row is not necessarily decisive. Overlapping primitives may cast redundant votes for the same class, or act on voxels whose winning class already leads by a wide margin. Conversely, a rarely active row can decide a small set of close-margin voxels. Most structured pruning methods also require a target row count or pruning ratio, and the right value varies substantially across architectures.

\textbf{Decision-FLIP} answers this question directly. Given a trained model and unlabeled calibration inputs, we remove each generated primitive from the occupancy readout alone and recompute the full class competition. The importance of a persistent bank row is the mean number of voxel decisions that change over the calibration frames. This score is output-aware, label-free, and tied to the prediction we want to preserve.

A ranking alone does not say how many rows to remove. Across architectures, log Mean-FLIP scores often show a concentrated low-impact group separated from a more influential tail, although we do not assume this is universal. We therefore treat budget selection as a two-population split and use Otsu's criterion to propose a data-dependent boundary without a target sparsity. Because Otsu always returns a threshold, a fixed separation gate rejects weakly supported splits and keeps the full bank instead. Section~\ref{sec:method} gives the criterion and the gate.

\begin{figure}[t]
\centering
\includegraphics[width=\linewidth]{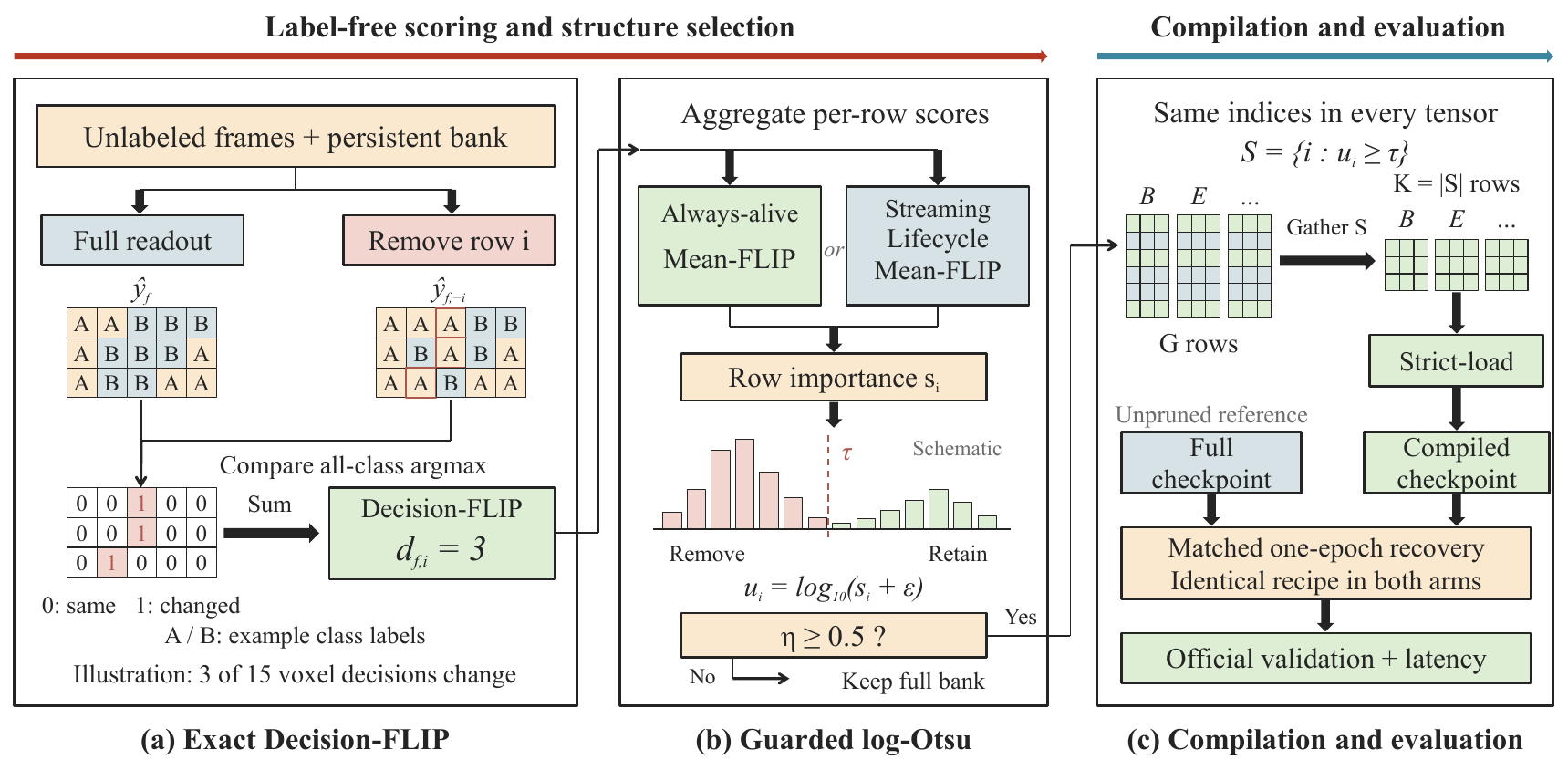}
\caption{Overview of Decision-FLIP. (a) On unlabeled calibration frames, removing row $i$ from the cached full readout and recomputing the all-class argmax gives per-frame Decision-FLIP; scores aggregate as Mean-FLIP or lifecycle Mean-FLIP. (b) Guarded log-Otsu selects a survivor set only when $\eta\geq0.5$, otherwise retaining the full bank. (c) The same survivor indices are gathered across aligned tensors, strict-loaded, and compared with the unpruned reference under matched one-epoch recovery and official evaluation.}
\label{fig:method-pipeline}
\end{figure}

Our contribution is not Otsu thresholding on its own, but an end-to-end framework that connects decision-level importance to physical model compilation (Figure~\ref{fig:method-pipeline}):

\begin{enumerate}[leftmargin=*]
\item \textbf{Decision-Level Scoring and Budget-Free Selection.} \emph{Decision-FLIP} measures primitive importance by exact voxel-decision changes, without labels, and pairs it with guarded Otsu selection to set the survivor count without a target sparsity.
\item \textbf{Lifecycle Integration and Physical Compilation.} We extend the criterion to streaming banks by following each persistent row through warp, death, and refill, and we compile the selected rows across all aligned tensors into strictly loadable checkpoints.
\item \textbf{Validation Across Models and Datasets.} Across four Gaussian-bank model-dataset settings and one persistent-query transfer, Decision-FLIP removes 12--79\% of eligible rows and reduces bank-dependent latency by up to 45\%, while the semantic difference after matched recovery stays within 0.18 point (Table~\ref{tab:main-summary}).
\end{enumerate}

\section{Related Work}

\subsection{Gaussian occupancy prediction}

GaussianFormer \citep{huang2024gaussianformer} represents scenes with sparse semantic Gaussians, while GaussianWorld \citep{zuo2025gaussianworld} extends Gaussian-space modeling to streaming occupancy. GaussianFormer3D~\citep{zhao2025gaussianformer3d} initializes Gaussians from LiDAR voxel features, and GaussianFlowOcc~\citep{boeder2025gaussianflowocc} predicts per-Gaussian temporal offsets. The released implementations of these models expose checkpoint-indexed primitive rows that meet the eligibility conditions in Section~\ref{sec:method}. OccFormer \citep{zhang2023occformer} exposes a checkpoint-indexed learned-query bank in the same way. GaussTR \citep{jiang2025gausstr}, in contrast, reuses 300 learned queries per view but resamples their 2D reference points at random on every forward pass, so a checkpoint row does not keep a stable primitive identity across calls.

\subsection{Output-aware pruning for occupancy}

Occupancy prediction combines a dominant empty class with dense all-class competition among overlapping primitives. A row can therefore be often active yet redundant for the decision, or rarely active but decisive. Whether local magnitude and activity proxies track the voxel decisions that compilation must preserve is thus an empirical question. Decision-FLIP measures exact voxel-decision changes under a readout intervention, and its offline compilation removes the selected rows from the checkpoint. This reduces row-dependent encoder, sampling, and readout computation, rather than filtering outputs only after inference. Efficient occupancy architectures attack related costs at the representation level. SparseOcc \citep{tang2024sparseocc} replaces dense 3D latent processing with a sparse latent volume and sparse operators, while OPUS \citep{wang2024opus} formulates occupancy as set prediction with learned queries instead of classifying a dense voxel grid. These methods redesign and train the predictor around a sparse representation; Decision-FLIP instead compiles redundant rows from an already trained, checkpoint-indexed bank.

\subsection{Output-aware compression and fair evaluation}

Structured pruning commonly uses magnitude or sensitivity criteria \citep{han2015learning,li2017pruning,molchanov2019importance}, while output-aware methods propagate final-response importance or preserve downstream features \citep{yu2018nisp,luo2017thinet,zhuang2018discrimination}. Exact Mean-FLIP directly measures nonlinear 0--1 voxel-decision changes instead of using a linearized proxy. Pruning structures have been selected using prescribed global thresholds \citep{liu2017slimming}, learned compression policies \citep{he2018amc}, or measured-latency constraints \citep{yang2018netadapt}. Controlled studies at matched architectures or sparsities \citep{liu2019rethinking,renda2020comparing} show that conclusions can depend strongly on the retraining protocol; we therefore give full and compiled models the same recovery budget. We apply Otsu's criterion \citep{otsu1979threshold} to log counterfactual scores behind a separation gate, producing a target-free candidate operating point without assuming universal bimodality.

\subsection{Dynamic token reduction and offline compilation}

DynamicViT \citep{rao2021dynamicvit} and Sparse DETR \citep{roh2022sparsedetr} select input-dependent token subsets, while ToMe \citep{bolya2023tome} merges input-dependent similar token pairs under a fixed per-layer merge count. These methods determine token reduction during inference and therefore retain runtime selection or similarity computation. Sparse DETR specifically sparsifies encoder-token updates according to expected decoder use; it does not prune a fixed persistent decoder-query bank. Decision-FLIP calibrates once and stores a physically reduced checkpoint, requiring no runtime selection policy. Per-scene Gaussian-compaction methods rank, mask, resample, or budget primitives optimized separately for each scene (LightGaussian \citep{fan2024lightgaussian}, Compact3DGS \citep{lee2024compact3dgs}, Mini-Splatting \citep{fang2024minisplatting}, and Taming 3DGS \citep{mallick2024taming}); our setting instead compiles checkpoint-indexed rows shared across inputs.

\section{Method}\label{sec:method}

\subsection{Persistent primitive banks}\label{sec:banks}

Let $B=\{b_i\}_{i=1}^{G}$ be a learned bank. For frame $f$, row $b_i$ generates a primitive $p_{f,i}$. The method requires:

\begin{itemize}[leftmargin=*]
\item a stable mapping from checkpoint row $i$ to generated primitive $p_{f,i}$;
\item an occupancy readout that can be recomputed after removing one primitive; and
\item aligned persistent tensors that can be gathered with the same survivor set.
\end{itemize}

We check these conditions in the implementation and, for streaming banks, in the explicit warp/death/refill lineage. Split-half stability tells us whether the calibration set is large enough; it does not establish row identity, and it cannot make an input-resampled row eligible. Appendix~\ref{sec:gausstr} shows what happens on GaussTR: the separation gate passes, yet the ranking is no better than random.

GaussianWorld and GaussianFormer-1 satisfy these requirements for their full banks. GaussianFormer3D, GaussianFlowOcc, and OccFormer also expose eligible persistent rows under the model-specific protocols in Appendix~\ref{sec:appendix-evidence}.

\subsection{Exact all-class removal}

Let $z_{f,c}(x)$ be the full class logit for voxel $x$, and let $\hat y_f(x)=\arg\max_c z_{f,c}(x)$. For an additive Gaussian readout,

\[
z_{f,c}(x)=e_c(x)+\sum_{i=1}^{G}q_{f,i,c}(x),
\]

where $q_{f,i,c}(x)$ is primitive $i$'s local contribution and $e_c(x)$ collects readout terms independent of the removable bank rows, including a base or empty-class term when present. Removing primitive $i$ is exact within the readout:

\[
z_{f,-i,c}(x)=z_{f,c}(x)-q_{f,i,c}(x).
\]

We recompute the argmax over \textbf{all classes}, including the empty class:

\[
\hat y_{f,-i}(x)=\arg\max_c z_{f,-i,c}(x).
\]

Additivity is required across cached row contributions, not within the upstream row generator: the query decoder and per-row heads may be nonlinear. For OccFormer, $q_{f,i,c}(x)$ is the post-softmax class probability multiplied by the post-sigmoid mask probability, and the final occupancy score sums these terms over queries. Removing one cached query contribution is therefore exact at the readout. Models whose readouts cannot be exactly decomposed or recomputed under row deletion are ineligible.

\subsection{Exact Mean-FLIP}

The per-row score is

\[
s_i=
\frac{1}{|\mathcal C|}
\sum_{f\in\mathcal C}
\sum_{x\in\mathcal V}
\mathbf 1\!\left[\hat y_f(x)\neq\hat y_{f,-i}(x)\right],
\]

where $\mathcal C$ contains fixed training-calibration frames and $\mathcal V$ is the voxel grid. Scoring uses no ground-truth labels. Equivalently, $s_i/|\mathcal V|$ is the empirical expected Hamming distance between the full model's voxel decisions and the single-row-intervened decisions. Exact Mean-FLIP is therefore an output-aware importance criterion that counts actual decision changes rather than approximating them.

Mean-FLIP measures a single-row readout intervention. It provides a ranking, but does not imply additivity or guarantee that a set of individually low-scoring rows can be removed jointly without decision drift. We therefore evaluate the complete selected mask before recovery and report the resulting zero-epoch operating point. A held-out set-removal ablation additionally measures joint drift against same-$K$ controls. Matched recovery adapts the physically reduced computation graph but is not treated as a guarantee of joint-removal safety.

To disambiguate, \textbf{Decision-FLIP} is the per-frame, per-row count of changed voxel decisions; \textbf{Mean-FLIP} is its calibration-set average; and \textbf{lifecycle Mean-FLIP} is the streaming aggregate that follows one persistent row's lineage through warp, death, and refill.

\subsection{Lifecycle Mean-FLIP aggregation}\label{sec:lifecycle}

For a non-recurrent bank, checkpoint row $i$ exists on every frame and the score above already averages its whole trajectory. Streaming models need a stricter notion of identity. A Gaussian can be warped into the next frame, leave the spatial range, die, or be replaced by a refill Gaussian. Treating the storage slot after a refill as the same primitive mixes different lifecycles and credits future evidence to the wrong checkpoint row.

We therefore follow each initial primitive through the model's exact warp validity mask. Let $\ell_i$ denote the initial Gaussian decoded from checkpoint row $i$, let $\mathcal L_i$ be the frames before that lineage leaves the valid range, and let $d_{f,\ell_i}$ be its exact all-class Decision-FLIP count on frame $f$. Its sequence-integrated importance is

\[
s_i^{\mathrm{life}}=
\frac{1}{|\mathcal C|}
\sum_{f\in\mathcal L_i} d_{f,\ell_i}.
\]

The denominator is the whole calibration horizon, not only the frames in which the Gaussian exists, so the score measures contribution over the full horizon rather than importance while alive. A short-lived primitive must therefore be highly decisive to outrank a consistently useful one, which means rare but consequential events can be underweighted. Death ends a lineage. GaussianWorld refill rows come from the decoder's fixed spatial prior rather than from a learned checkpoint row, so they start new, unscored lineages and carry no evidence across that boundary. Compilation sets both the initial learned-bank size and the decoder's live-state budget to the selected $K$. For always-alive banks (GaussianFormer-1, GaussianFormer3D, GaussianFlowOcc, and OccFormer), $\mathcal L_i=\mathcal C$ and $s_i^{\mathrm{life}}$ equals Mean-FLIP.

\subsection{Guarded Otsu budget selection}\label{sec:otsu}

We transform the scores as

\[
u_i=\log_{10}(s_i+\epsilon)
\]

and choose the Otsu threshold $\tau$ that maximizes the between-group variance. The offset $\epsilon$ is fixed before thresholding to $10^{-3}$ times the first percentile of the positive scores ($10^{-12}$ if every score is zero). Its only role is to place exact zeros on the log axis; it lies below every positive score and does not change their order. The normalized separation quality of the split is

\[
\eta=\frac{\sigma^2_{\mathrm{between}}}{\sigma^2_{\mathrm{total}}}.
\]

One configuration is fixed for the whole study, with no model- or dataset-specific tuning:

\begin{itemize}[leftmargin=*]
\item minimum separation: $\eta_{\min}=0.5$;
\item no target $K$ or fixed removal fraction; and
\item abstain rather than invent a cut when the separation gate fails.
\end{itemize}

If the gate passes, we keep rows with $u_i\ge\tau$ and remove the rest; ties at the threshold are kept. If it fails, the selector abstains and returns the original bank ($K=G$). The survivor count is thus read off the score distribution rather than supplied. The gate decides only whether to accept or abstain; once it accepts, the Otsu threshold alone determines $K$. \textbf{Budget-free} therefore means that no survivor count or pruning ratio is given, not that the protocol has no constants: $\epsilon$ and $\eta_{\min}$ are fixed. The gate checks score separation, not statistical confidence; split-half stability and the full selected mask are evaluated separately.

\subsection{Physical compilation and matched evaluation}

The survivor indices are gathered across every aligned persistent tensor, and the reduced checkpoint must load under strict shape and key checking. Scoring and selection end here and use no labels. We report the compiled checkpoint before recovery, and also after an optional one-epoch recovery that lets row-dependent sampling, attention, and normalization adapt to the smaller tensors. For a fair comparison, the full and compiled models receive the same recovery recipe while the survivor mask and compiled shape stay fixed. Recovery trains every parameter of the compiled checkpoint, including the surviving bank rows, unless an experiment states that a module is frozen, and it uses task labels unless marked self-distillation-only.

\section{Experiments}\label{sec:experiments}

\subsection{Protocol}

We evaluate five eligible model-dataset settings. Four use Gaussian banks: GaussianWorld, GaussianFormer-1, and GaussianFormer3D on nuScenes-SurroundOcc, and GaussianFlowOcc on Occ3D-nuScenes~\citep{tian2023occ3d}. The fifth, OccFormer on SemanticKITTI, tests transfer to a non-Gaussian persistent-query bank. We use each benchmark's official split and evaluator where possible \citep{wei2023surroundocc,caesar2020nuscenes,tian2023occ3d,behley2019semantickitti}. GaussianFormer3D uses keyframe LiDAR only in every arm, and GaussianFlowOcc is our 3D-supervised reproduction; both are compared only within architecture. GaussTR is a negative control without stable row identity (Appendix~\ref{sec:gausstr}).

Scores are computed on unlabeled training-calibration inputs. Guarded Otsu with the fixed $\eta_{\min}=0.5$ gate chooses the removal count, survivor indices are gathered across every aligned tensor, and the reduced checkpoints load under strict shape and key checking. Every primary comparison gives the full and compiled models the same one-epoch recovery budget; same-$K$ ranking controls and GaussTR receive no recovery. The extra epoch alone raises GaussianFormer-1 and GaussianFormer3D by about 1.7 and 2.7 mIoU, so matched full controls are essential. Evaluations use each complete split, including GaussianWorld's 5,700-frame streaming protocol. Latency is measured in PyTorch fp32 at batch size 1 on one NVIDIA L4, with both configurations resident and interleaved after warm-up, except where stated; bank/query-dependent and whole-model changes are reported separately. Peak GPU memory is measured for the full and compiled shapes in separate fresh processes. Calibration sizes, evaluator details, matched-recovery results, checkpoint fidelity, and timing breakdowns are in Appendix~\ref{sec:appendix-evidence}.

\subsection{Results across models and datasets}

Table~\ref{tab:main-summary} combines the automatically selected structure, matched accuracy, paired latency, and peak-memory result for every eligible model-dataset setting, separating the Gaussian-bank study from the SemanticKITTI persistent-query transfer. The semantic difference after matched recovery is at most 0.171 point across the primary adaptive rows, while automatic removal ranges from 12\% to 79\% of the eligible bank (minimum 12.0\% on GaussianFormer3D with nuScenes-SurroundOcc; maximum 79.0\% on OccFormer). Peak-memory savings reach 16.6\%, although fixed activations make the change negligible on several models and GaussianFormer-1 increases by 0.8\%. For the five primary pairs with complete per-class outputs, no class that is nonzero in its matched full model newly collapses to zero after adaptive compilation.

\begin{table}[t]
\caption{Automatic compilation across model-dataset settings. Changes are compiled minus full in the displayed metric order.}\label{tab:main-summary}
\centering
\footnotesize
\setlength{\tabcolsep}{2pt}
\begin{tabular}{>{\raggedright\arraybackslash}p{0.19\linewidth}>{\raggedleft\arraybackslash}p{0.14\linewidth}>{\raggedleft\arraybackslash}p{0.13\linewidth}>{\raggedleft\arraybackslash}p{0.15\linewidth}>{\raggedleft\arraybackslash}p{0.09\linewidth}>{\raggedleft\arraybackslash}p{0.09\linewidth}>{\raggedleft\arraybackslash}p{0.08\linewidth}}
\toprule
\textbf{Model / metrics} & \textbf{Kept / total (removed)} & \textbf{Before $\Delta$} & \textbf{After recovery $\Delta$} & \textbf{Bank/\allowbreak query latency $\Delta$} & \textbf{End-to-end latency $\Delta$} & \textbf{Peak memory $\Delta$} \\
\midrule
GaussianWorld, primary lifecycle compilation (mIoU / geometry IoU) & 22,427 / 25,600 (12.4\%) & -0.179 / -0.290 & -0.082 / -0.119 & -7.0\% & -2.4\% & -1.2\% \\
\emph{GaussianWorld, non-primary static-frame reference} (mIoU / geometry IoU) & \emph{16,461 / 25,600 (35.7\%)} & \emph{-0.412 / -0.574} & \emph{-0.225 / -0.270} & \emph{-24.3\%} & \emph{-9.3\%} & \emph{0.0\%} \\
GaussianFormer-1 (mIoU / geometry IoU) & 17,365 / 25,600 (32.2\%) & -0.157 / -0.435 & -0.171 / -0.246 & -23.3\% & -5.5\% & +0.8\% \\
GaussianFormer3D (mIoU / geometry IoU) & 22,535 / 25,600 (12.0\%) & -0.044 / -0.093 & -0.024 / -0.116 & -13.2\% & -4.7\% & -8.3\% \\
GaussianFlowOcc (voxel mIoU / occupied IoU) & 8,628 / 10,000 (13.72\%) & +0.10 / +0.57 & +0.01 / +0.03 & -10.7\% & -8.0\% & -7.0\% \\
\midrule
OccFormer transfer, SemanticKITTI (SSC mIoU / SC IoU) & 21 / 100 (79.0\%) & -0.16 / 0.00 & +0.03 / +0.08 & -45.0\% & -5.7\% & -16.6\% \\
\bottomrule
\end{tabular}%
\par\vspace{3pt}\begin{minipage}{\linewidth}\footnotesize Deltas are compiled minus full at the same stage, and italics mark the non-primary GaussianWorld static-frame reference. Peak memory is fresh-process PyTorch peak allocated memory after warm-up and reset for every row; reserved and NVML process measurements are diagnostics only (Appendix~\ref{sec:protocols}).\end{minipage}
\end{table}

For GaussianWorld, the lifecycle row is the primary result: it follows persistent-row identity through warp, death, and refill over the streaming sequence. The static-frame reference instead scores checkpoint rows independently per frame and ignores lineage events; it selects a more aggressive $K$ and is reported only as a separate deployment trade-off whose values must not be combined with the lifecycle row. The lifecycle compilation costs 0.082 mIoU while cutting whole-model latency by 2.4\%. Across settings with component profiles, the slot-dependent kernels speed up considerably, but the image backbone, voxel encoder or decoder, or LiDAR branch runs unchanged, and Amdahl's law caps the end-to-end gain. The 7.0--45.0\% bank/query-dependent reductions verify that physical compilation removes the intended row-dependent work; they are mechanism measurements, not substitutes for the paired 2.4--8.0\% primary whole-model results. Memory effects are likewise architecture dependent: OccFormer saves 16.6\% peak allocated memory and GaussianFlowOcc saves 7.0\%, whereas GaussianFormer-1 is unchanged within measurement variation (+0.8\%). We therefore report component, whole-model, and memory results together rather than extrapolating unmeasured gains to other deployment hardware.

\begin{figure}[t]
\centering
\includegraphics[width=\linewidth]{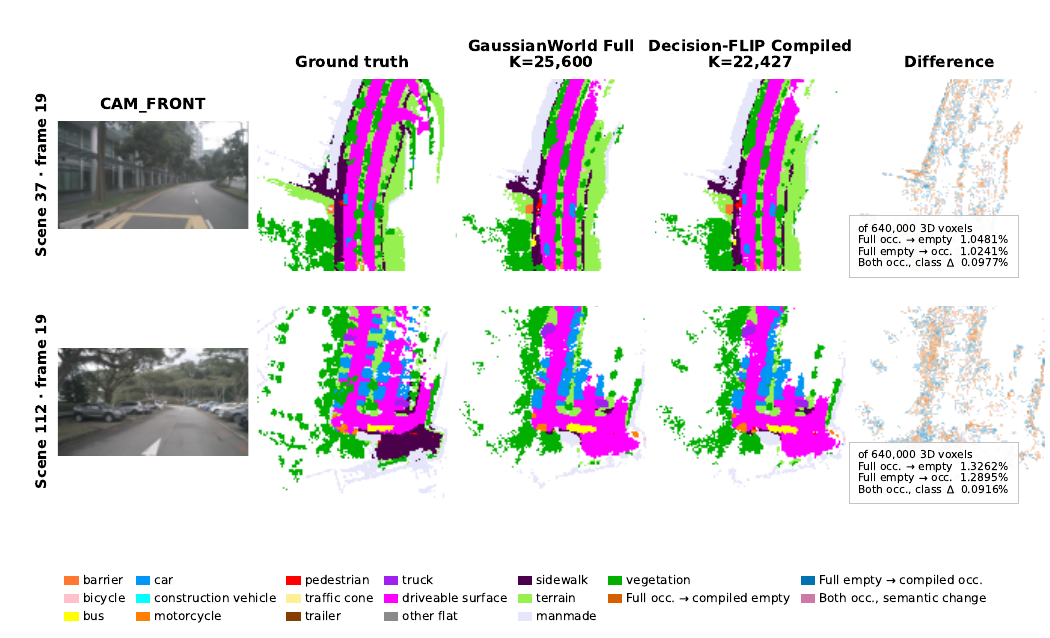}
\caption{Qualitative preservation under lifecycle-aware GaussianWorld compilation. The two validation samples (scene indices 37 and 112, frame 19) were fixed before inspecting predictions. Each model processes the same 20-frame streaming prefix after matched recovery. Decision-FLIP retains 22,427 of 25,600 rows (12.4\% removed). Difference panels separate full occupied to compiled empty, full empty to compiled occupied, and semantic changes where both predictions remain occupied; percentages use all 640,000 voxels in each 200$\times$200$\times$16 grid.}
\label{fig:gw-qualitative}
\end{figure}

Figure~\ref{fig:gw-qualitative} compares the primary lifecycle-compiled GaussianWorld model with its matched full control on two fixed validation samples. Scene geometry and semantic layout are closely preserved in both. Most differences lie on occupancy boundaries; semantic relabeling among voxels that stay occupied accounts for less than 0.10\% of the voxel grid in either sample. Qualitative samples for the remaining architectures, fixed before any output was inspected, are in Appendix~\ref{sec:qualitative-results}.

\begin{figure}[t]
\centering
\includegraphics[width=\linewidth]{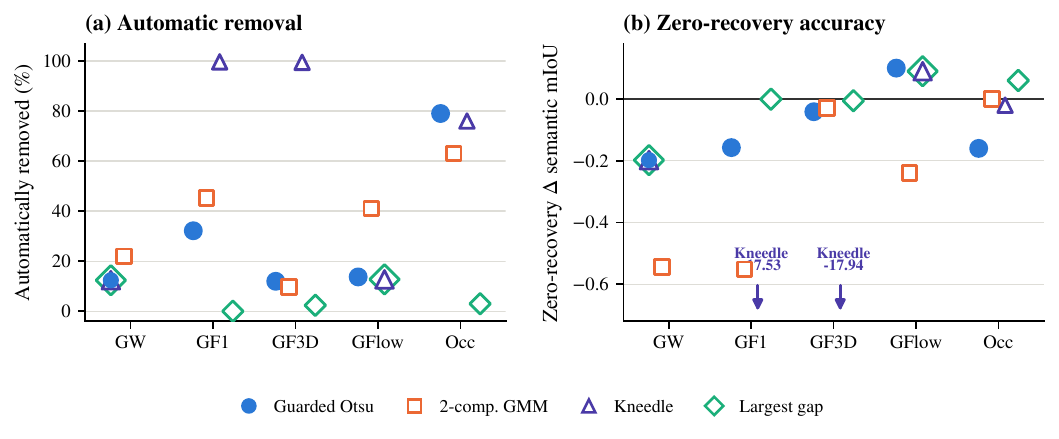}
\caption{Automatic selector comparison. (a) Automatically removed fraction. (b) Zero-recovery semantic mIoU change from each model's full source. Concentric markers indicate identical masks; arrows annotate clipped Kneedle failures. Selection never uses validation metrics.}
\label{fig:selector-ablation}
\end{figure}

Figure~\ref{fig:selector-ablation} compares guarded Otsu with Kneedle \citep{satopaa2011kneedle}, a two-component Gaussian mixture model (GMM) \citep{mclachlan2019mixture}, and a largest-gap rule (cut at the widest gap between sorted log scores) under a common zero-recovery protocol; these are baselines, and guarded Otsu remains the fixed selector for all primary comparisons. Guarded Otsu selects non-trivial cuts of 12--79\% on all five settings. GMM removes more rows but incurs larger losses on GaussianWorld, GaussianFormer-1, and GaussianFlowOcc; Kneedle collapses GaussianFormer-1 and GaussianFormer3D; and largest-gap often removes almost nothing. Some alternatives match or slightly exceed Otsu on individual architectures. This supports guarded Otsu as a non-degenerate, Pareto-competitive operating point, not as universally accuracy-optimal. On GaussianFlowOcc, guarded Otsu retains 8,628 rows, whereas Kneedle and largest-gap share the distinct 8,725-row mask. Appendix~\ref{sec:selector-comparison} gives the complete operating points and the recovered GaussianFormer-1 analysis.

\subsection{Same-\texorpdfstring{$K$}{K} ranking-criterion ablation}

To separate ranking quality from supervised re-adaptation, Table~\ref{tab:main-samek} changes only the ranking at each automatically selected survivor count, applies no recovery, and evaluates the complete validation split from the same source checkpoint. Mean-FLIP and the heuristic controls use unlabeled calibration frames; Taylor \citep{molchanov2019importance} uses training labels only to form its ranking. To fit every architecture in the main paper, entries are changes in the primary semantic metric relative to the same full source checkpoint (mIoU, or SSC mIoU for OccFormer); full metric pairs and absolute values are in Appendix~\ref{sec:appendix-evidence}. Random removal collapses GaussianFormer-1 and OccFormer, and on OccFormer the fire-rate proxy erases three rare classes (bicycle, motorcycle, traffic-sign) that Mean-FLIP keeps. At the same $K$, label-free Mean-FLIP matches or outperforms supervised Taylor on all five settings, by a wide margin on GaussianFormer-1 and OccFormer. This supports the decision-level ranking signal; it is not a guarantee about joint removal. The selector comparison above addresses budget selection separately from these ranking controls.

\begin{table}[t]
\caption{Same-$K$ ranking-criterion ablation before recovery (primary semantic metric $\Delta$ vs. full). We report one primary automatically selected operating point per architecture; the non-primary GaussianWorld static reference is excluded.}\label{tab:main-samek}
\centering
\small
\setlength{\tabcolsep}{3pt}
\begin{tabular}{>{\raggedright\arraybackslash}p{0.18\linewidth}>{\raggedleft\arraybackslash}p{0.06\linewidth}>{\raggedleft\arraybackslash}p{0.10\linewidth}>{\raggedleft\arraybackslash}p{0.09\linewidth}>{\raggedleft\arraybackslash}p{0.09\linewidth}>{\raggedleft\arraybackslash}p{0.09\linewidth}>{\raggedleft\arraybackslash}p{0.09\linewidth}>{\raggedleft\arraybackslash}p{0.12\linewidth}}
\toprule
\textbf{Model} & \textbf{$K$} & \textbf{Mean-FLIP} & \textbf{Taylor} & \textbf{Random} & \textbf{Activity} & \textbf{Opacity} & \textbf{Magnitude} \\
\midrule
GaussianWorld & 22,427 & \textbf{-0.179} & -0.194 & -0.390 & -0.252 & -0.244 & -0.326 \\
GaussianFormer-1 & 17,365 & \textbf{-0.157} & -1.221 & -8.310 & -8.652 & -0.183 & -0.474 \\
GaussianFormer3D & 22,535 & \textbf{-0.044} & -0.075 & -0.730 & -0.100 & -0.105 & -0.071 \\
GaussianFlowOcc & 8,628 & \textbf{+0.10} & \textbf{+0.10} & n/a & n/a & n/a & n/a \\
\midrule
OccFormer (SemanticKITTI transfer) & 21 & \textbf{-0.16} & -7.98 & -12.68 & -0.33 & n/a & -11.97 \\
\bottomrule
\end{tabular}%
\par\vspace{3pt}\begin{minipage}{\linewidth}\footnotesize \textbf{Proxy definitions.} Taylor uses training labels and ranks rows by the calibration mean of $\sum_{p\in i}\lvert w_p\,\partial\mathcal L/\partial w_p\rvert$ over each aligned row $i$. Random uses seed 0; activity and opacity are calibration statistics; magnitude/query norm is each implementation's non-counterfactual parameter-magnitude proxy. All arms use zero recovery and the same $K$ within each model. Bold marks the best result in each row; \texttt{\detokenize{n/a}} marks a proxy that is undefined for the architecture or was not run at this $K$.\end{minipage}
\end{table}

\subsection{Per-class preservation}

Aggregate mIoU can hide failures on rare categories, so complete per-class outputs and a worst-change summary are in Appendix~\ref{sec:perclass-results}. Among the five primary matched pairs with complete per-class outputs, no class that is nonzero in its matched full model newly collapses to zero, and the largest class-level decrease is 0.59 points (OccFormer other-vehicle).

\section{Discussion}

Decision-FLIP applies when rows are identifiable, exactly removable at the all-class readout, and physically compilable across aligned tensors. OccFormer shows that these conditions extend beyond Gaussian banks to persistent learned queries, and GaussianWorld shows why streaming state needs lifecycle-aware scoring. The experiments support effective ranking and architecture-specific budget selection, not universal dominance or statistical equivalence; only GaussianFormer-1 was run with three recovery seeds, and the other settings use one. Same-$K$ controls isolate ranking quality, and zero-recovery results show that the signal is present before any fine-tuning. Finally, fixed-cost image, voxel, and sensor modules can limit end-to-end acceleration even when bank-dependent computation falls substantially.

\section{Limitations}

Decision-FLIP measures changes relative to a trained model's readout, so it cannot tell correct predictions from teacher errors. Scoring reuses cached upstream outputs rather than rerunning row-dependent sampling, attention, and feature interaction for every deletion. Individually low scores therefore do not guarantee drift-free joint removal of correlated rows, and frame-averaged scores may underweight rare but consequential events. The method needs a stable mapping from checkpoint rows to generated primitives, which rules out architectures that resample or reorder query slots on every input. Streaming models must follow spatial warps and end lineages explicitly at death and refill boundaries. Optional recovery adds training cost and normally uses labels. Finally, whole-model speedups remain bounded by fixed-cost image, voxel, or sensor modules, so absolute and relative savings depend on the target hardware and kernels.

\section{Conclusion}

Decision-FLIP compiles persistent occupancy banks by ranking rows with unlabeled all-class readout interventions and selecting $K$ through guarded Otsu. Across four Gaussian model-dataset settings and one persistent-query transfer, it removes 12--79\% of eligible rows, reduces row-dependent latency by up to 45\% and whole-model latency by up to 8.0\% on the primary endpoints, and keeps the primary matched-recovery semantic changes within 0.18 point. Same-$K$ controls confirm that decision-aware ranking matters. These results establish compilation without a target sparsity for identifiable persistent rows, with end-to-end gains bounded by fixed-cost modules.

\begingroup
\small
\subsection*{Ethics statement}
This work studies compression of 3D occupancy prediction systems using public
autonomous-driving benchmarks. It introduces no human-subject data. As with
other perception compression methods, deployment requires validation under the
target sensor, hardware, and safety conditions; the reported benchmark results
do not by themselves establish safety for real-world driving.

\subsection*{Reproducibility statement}
Sections~\ref{sec:method} and~\ref{sec:experiments} specify the scoring,
lifecycle, thresholding, compilation,
matched-recovery, and evaluation protocols. The appendices report recovery
controls and temporal diagnostics. Code, fixed checkpoint hashes, calibration
splits, survivor counts, complete-validation metrics, and strict-loading
checks will be released upon acceptance.

\subsection*{AI use statement}
Generative AI tools were used to assist with language editing and \LaTeX{}
formatting. The authors
take full responsibility for the final content of this work, including all
AI-assisted text.
\endgroup

\bibliography{iclr2027_conference}

@InProceedings{huang2024gaussianformer,
author="Huang, Yuanhui
and Zheng, Wenzhao
and Zhang, Yunpeng
and Zhou, Jie
and Lu, Jiwen",
editor="Leonardis, Ale{\v{s}}
and Ricci, Elisa
and Roth, Stefan
and Russakovsky, Olga
and Sattler, Torsten
and Varol, G{\"u}l",
title="GaussianFormer: Scene as Gaussians for Vision-Based 3D Semantic Occupancy Prediction",
booktitle="Computer Vision -- ECCV 2024",
year="2025",
publisher="Springer Nature Switzerland",
address="Cham",
pages="376--393",
isbn="978-3-031-73383-3"
}

@INPROCEEDINGS{zuo2025gaussianworld,
  author={Zuo, Sicheng and Zheng, Wenzhao and Huang, Yuanhui and Zhou, Jie and Lu, Jiwen},
  booktitle={2025 IEEE/CVF Conference on Computer Vision and Pattern Recognition (CVPR)}, 
  title={GaussianWorld: Gaussian World Model for Streaming 3D Occupancy Prediction}, 
  year={2025},
  volume={},
  number={},
  pages={6772-6781},
  doi={10.1109/CVPR52734.2025.00635}}

@article{otsu1979threshold,
  title={A threshold selection method from gray-level histograms},
  author={Otsu, Nobuyuki},
  journal={IEEE transactions on systems, man, and cybernetics},
  volume={9},
  number={1},
  pages={62--66},
  year={1979},
  publisher={IEEE},
  doi     = {10.1109/TSMC.1979.4310076},
  url     = {https://ieeexplore.ieee.org/document/4310076}
}

@INPROCEEDINGS{wei2023surroundocc,
  author={Wei, Yi and Zhao, Linqing and Zheng, Wenzhao and Zhu, Zheng and Zhou, Jie and Lu, Jiwen},
  booktitle={2023 IEEE/CVF International Conference on Computer Vision (ICCV)}, 
  title={SurroundOcc: Multi-Camera 3D Occupancy Prediction for Autonomous Driving}, 
  year={2023},
  volume={},
  number={},
  pages={21672-21683},
  doi={10.1109/ICCV51070.2023.01986}}

@INPROCEEDINGS{caesar2020nuscenes,
  author={Caesar, Holger and Bankiti, Varun and Lang, Alex H. and Vora, Sourabh and Liong, Venice Erin and Xu, Qiang and Krishnan, Anush and Pan, Yu and Baldan, Giancarlo and Beijbom, Oscar},
  booktitle={2020 IEEE/CVF Conference on Computer Vision and Pattern Recognition (CVPR)}, 
  title={nuScenes: A Multimodal Dataset for Autonomous Driving}, 
  year={2020},
  volume={},
  number={},
  pages={11618-11628},
  doi={10.1109/CVPR42600.2020.01164}}

@inproceedings{han2015learning,
 author = {Han, Song and Pool, Jeff and Tran, John and Dally, William},
 booktitle = {Advances in Neural Information Processing Systems},
 editor = {C. Cortes and N. Lawrence and D. Lee and M. Sugiyama and R. Garnett},
 pages = {},
 publisher = {Curran Associates, Inc.},
 title = {Learning both Weights and Connections for Efficient Neural Network},
 url = {https://proceedings.neurips.cc/paper_files/paper/2015/file/ae0eb3eed39d2bcef4622b2499a05fe6-Paper.pdf},
 volume = {28},
 year = {2015}
}

@inproceedings{li2017pruning,
  title     = {Pruning Filters for Efficient {ConvNets}},
  author    = {Li, Hao and Kadav, Asim and Durdanovic, Igor and Samet, Hanan and Graf, Hans Peter},
  booktitle = {International Conference on Learning Representations},
  year      = {2017},
  url       = {https://openreview.net/forum?id=rJqFGTslg}
}

@INPROCEEDINGS{molchanov2019importance,
  author={Molchanov, Pavlo and Mallya, Arun and Tyree, Stephen and Frosio, Iuri and Kautz, Jan},
  booktitle={2019 IEEE/CVF Conference on Computer Vision and Pattern Recognition (CVPR)}, 
  title={Importance Estimation for Neural Network Pruning}, 
  year={2019},
  volume={},
  number={},
  pages={11256-11264},
  doi={10.1109/CVPR.2019.01152}}

@inproceedings{liu2019rethinking,
  title     = {Rethinking the Value of Network Pruning},
  author    = {Liu, Zhuang and Sun, Mingjie and Zhou, Tinghui and Huang, Gao and Darrell, Trevor},
  booktitle = {International Conference on Learning Representations},
  year      = {2019},
  url       = {https://openreview.net/forum?id=rJlnB3C5Ym}
}

@inproceedings{renda2020comparing,
  title     = {Comparing Rewinding and Fine-tuning in Neural Network Pruning},
  author    = {Renda, Alex and Frankle, Jonathan and Carbin, Michael},
  booktitle = {International Conference on Learning Representations},
  year      = {2020},
  url       = {https://openreview.net/forum?id=S1gSj0NKvB}
}

@INPROCEEDINGS{yu2018nisp,
  author={Yu, Ruichi and Li, Ang and Chen, Chun-Fu and Lai, Jui-Hsin and Morariu, Vlad I. and Han, Xintong and Gao, Mingfei and Lin, Ching-Yung and Davis, Larry S.},
  booktitle={2018 IEEE/CVF Conference on Computer Vision and Pattern Recognition}, 
  title={NISP: Pruning Networks Using Neuron Importance Score Propagation}, 
  year={2018},
  volume={},
  number={},
  pages={9194-9203},
  doi={10.1109/CVPR.2018.00958}}

@INPROCEEDINGS{luo2017thinet,
  author={Luo, Jian-Hao and Wu, Jianxin and Lin, Weiyao},
  booktitle={2017 IEEE International Conference on Computer Vision (ICCV)}, 
  title={ThiNet: A Filter Level Pruning Method for Deep Neural Network Compression}, 
  year={2017},
  volume={},
  number={},
  pages={5068-5076},
  doi={10.1109/ICCV.2017.541}}

@inproceedings{zhuang2018discrimination,
 author = {Zhuang, Zhuangwei and Tan, Mingkui and Zhuang, Bohan and Liu, Jing and Guo, Yong and Wu, Qingyao and Huang, Junzhou and Zhu, Jinhui},
 booktitle = {Advances in Neural Information Processing Systems},
 editor = {S. Bengio and H. Wallach and H. Larochelle and K. Grauman and N. Cesa-Bianchi and R. Garnett},
 pages = {},
 publisher = {Curran Associates, Inc.},
 title = {Discrimination-aware Channel Pruning for Deep Neural Networks},
 url = {https://proceedings.neurips.cc/paper_files/paper/2018/file/55a7cf9c71f1c9c495413f934dd1a158-Paper.pdf},
 volume = {31},
 year = {2018}
}

@INPROCEEDINGS{liu2017slimming,
  author={Liu, Zhuang and Li, Jianguo and Shen, Zhiqiang and Huang, Gao and Yan, Shoumeng and Zhang, Changshui},
  booktitle={2017 IEEE International Conference on Computer Vision (ICCV)}, 
  title={Learning Efficient Convolutional Networks through Network Slimming}, 
  year={2017},
  volume={},
  number={},
  pages={2755-2763},
  doi={10.1109/ICCV.2017.298}}

@InProceedings{he2018amc,
author="He, Yihui
and Lin, Ji
and Liu, Zhijian
and Wang, Hanrui
and Li, Li-Jia
and Han, Song",
editor="Ferrari, Vittorio
and Hebert, Martial
and Sminchisescu, Cristian
and Weiss, Yair",
title="AMC: AutoML for Model Compression and Acceleration on Mobile Devices",
booktitle="Computer Vision -- ECCV 2018",
year="2018",
publisher="Springer International Publishing",
address="Cham",
pages="815--832",
isbn="978-3-030-01234-2"
}

@InProceedings{yang2018netadapt,
author="Yang, Tien-Ju
and Howard, Andrew
and Chen, Bo
and Zhang, Xiao
and Go, Alec
and Sandler, Mark
and Sze, Vivienne
and Adam, Hartwig",
editor="Ferrari, Vittorio
and Hebert, Martial
and Sminchisescu, Cristian
and Weiss, Yair",
title="NetAdapt: Platform-Aware Neural Network Adaptation for Mobile Applications",
booktitle="Computer Vision -- ECCV 2018",
year="2018",
publisher="Springer International Publishing",
address="Cham",
pages="289--304",
isbn="978-3-030-01249-6"
}

@inproceedings{fan2024lightgaussian,
 author = {Fan, Zhiwen and Wang, Kevin and Wen, Kairun and Zhu, Zehao and Xu, Dejia and Wang, Zhangyang},
 booktitle = {Advances in Neural Information Processing Systems},
 doi = {10.52202/079017-4447},
 editor = {A. Globerson and L. Mackey and D. Belgrave and A. Fan and U. Paquet and J. Tomczak and C. Zhang},
 pages = {140138--140158},
 publisher = {Curran Associates, Inc.},
 title = {LightGaussian: Unbounded 3D Gaussian Compression with 15x Reduction and 200+ FPS},
 url = {https://proceedings.neurips.cc/paper_files/paper/2024/file/fd881d3b625437354d4421818f81058f-Paper-Conference.pdf},
 volume = {37},
 year = {2024}
}

@INPROCEEDINGS{lee2024compact3dgs,
  author={Lee, Joo Chan and Rho, Daniel and Sun, Xiangyu and Ko, Jong Hwan and Park, Eunbyung},
  booktitle={2024 IEEE/CVF Conference on Computer Vision and Pattern Recognition (CVPR)}, 
  title={Compact 3D Gaussian Representation for Radiance Field}, 
  year={2024},
  volume={},
  number={},
  pages={21719-21728},
  doi={10.1109/CVPR52733.2024.02052}}

@inproceedings{fang2024minisplatting,
  title     = {{Mini-Splatting}: Representing Scenes with a Constrained Number of {Gaussians}},
  author    = {Fang, Guangchi and Wang, Bing},
  booktitle = {Computer Vision -- ECCV 2024},
  series    = {Lecture Notes in Computer Science},
  volume    = {15135},
  pages     = {165--181},
  year      = {2024},
  doi       = {10.1007/978-3-031-72980-5_10},
  url       = {https://doi.org/10.1007/978-3-031-72980-5_10}
}

@inproceedings{mallick2024taming,
  title     = {Taming {3DGS}: High-Quality Radiance Fields with Limited Resources},
  author    = {Mallick, Saswat Subhajyoti and Goel, Rahul and Kerbl, Bernhard and Steinberger, Markus and Carrasco, Francisco Vicente and De La Torre, Fernando},
  booktitle = {SIGGRAPH Asia 2024 Conference Papers},
  articleno = {2},
  numpages  = {11},
  year      = {2024},
  doi       = {10.1145/3680528.3687694},
  url       = {https://dl.acm.org/doi/10.1145/3680528.3687694}
}

@inproceedings{rao2021dynamicvit,
  title     = {{DynamicViT}: Efficient Vision Transformers with Dynamic Token Sparsification},
  author    = {Rao, Yongming and Zhao, Wenliang and Liu, Benlin and Lu, Jiwen and Zhou, Jie and Hsieh, Cho-Jui},
  booktitle = {Advances in Neural Information Processing Systems},
  volume    = {34},
  pages     = {13937--13949},
  year      = {2021},
  url       = {https://proceedings.neurips.cc/paper/2021/hash/747d3443e319a22747fbb873e8b2f9f2-Abstract.html}
}

@inproceedings{roh2022sparsedetr,
  title     = {Sparse {DETR}: Efficient End-to-End Object Detection with Learnable Sparsity},
  author    = {Roh, Byungseok and Shin, JaeWoong and Shin, Wuhyun and Kim, Saehoon},
  booktitle = {International Conference on Learning Representations},
  year      = {2022},
  url       = {https://openreview.net/forum?id=RRGVCN8kjim}
}

@inproceedings{bolya2023tome,
  title     = {Token Merging: Your {ViT} But Faster},
  author    = {Bolya, Daniel and Fu, Cheng-Yang and Dai, Xiaoliang and Zhang, Peizhao and Feichtenhofer, Christoph and Hoffman, Judy},
  booktitle = {International Conference on Learning Representations},
  year      = {2023},
  url       = {https://openreview.net/forum?id=JroZRaRw7Eu}
}

@inproceedings{zhang2023occformer,
  title     = {{OccFormer}: Dual-path Transformer for Vision-based {3D} Semantic Occupancy Prediction},
  author    = {Zhang, Yunpeng and Zhu, Zheng and Du, Dalong},
  booktitle = {Proceedings of the IEEE/CVF International Conference on Computer Vision},
  pages     = {9433--9443},
  year      = {2023},
  doi       = {10.1109/ICCV51070.2023.00865},
  url       = {https://openaccess.thecvf.com/content/ICCV2023/html/Zhang_OccFormer_Dual-path_Transformer_for_Vision-based_3D_Semantic_Occupancy_Prediction_ICCV_2023_paper.html}
}

@article{zhao2025gaussianformer3d,
  title         = {{GaussianFormer3D}: Multi-Modal {Gaussian}-based Semantic Occupancy Prediction with {3D} Deformable Attention},
  author        = {Zhao, Lingjun and Wei, Sizhe and Hays, James and Gan, Lu},
  journal       = {arXiv preprint arXiv:2505.10685},
  year          = {2025},
  eprint        = {2505.10685},
  archiveprefix = {arXiv},
  note          = {Accepted to IEEE ICRA 2026},
  url           = {https://arxiv.org/abs/2505.10685}
}

@inproceedings{boeder2025gaussianflowocc,
  title     = {{GaussianFlowOcc}: Sparse and Weakly Supervised Occupancy Estimation using {Gaussian} Splatting and Temporal Flow},
  author    = {Boeder, Simon and Gigengack, Fabian and Risse, Benjamin},
  booktitle = {Proceedings of the IEEE/CVF International Conference on Computer Vision},
  pages     = {24943--24954},
  year      = {2025},
  doi       = {10.1109/ICCV51701.2025.02313},
  url       = {https://openaccess.thecvf.com/content/ICCV2025/html/Boeder_GaussianFlowOcc_Sparse_and_Weakly_Supervised_Occupancy_Estimation_using_Gaussian_Splatting_ICCV_2025_paper.html}
}

@inproceedings{tian2023occ3d,
  title     = {{Occ3D}: A Large-Scale {3D} Occupancy Prediction Benchmark for Autonomous Driving},
  author    = {Tian, Xiaoyu and Jiang, Tao and Yun, Longfei and Mao, Yucheng and Yang, Huitong and Wang, Yue and Wang, Yilun and Zhao, Hang},
  booktitle = {Advances in Neural Information Processing Systems},
  volume    = {36},
  pages     = {64318--64330},
  year      = {2023},
  note      = {Datasets and Benchmarks Track},
  url       = {https://proceedings.neurips.cc/paper_files/paper/2023/hash/cabfaeecaae7d6540ee797a66f0130b0-Abstract-Datasets_and_Benchmarks.html}
}

@inproceedings{behley2019semantickitti,
  title     = {{SemanticKITTI}: A Dataset for Semantic Scene Understanding of {LiDAR} Sequences},
  author    = {Behley, Jens and Garbade, Martin and Milioto, Andres and Quenzel, Jan and Behnke, Sven and Stachniss, Cyrill and Gall, J{\"u}rgen},
  booktitle = {Proceedings of the IEEE/CVF International Conference on Computer Vision},
  pages     = {9297--9307},
  year      = {2019},
  doi       = {10.1109/ICCV.2019.00939},
  url       = {https://openaccess.thecvf.com/content_ICCV_2019/html/Behley_SemanticKITTI_A_Dataset_for_Semantic_Scene_Understanding_of_LiDAR_Sequences_ICCV_2019_paper.html}
}

@inproceedings{jiang2025gausstr,
  title     = {{GaussTR}: Foundation Model-Aligned {Gaussian} Transformer for Self-Supervised {3D} Spatial Understanding},
  author    = {Jiang, Haoyi and Liu, Liu and Cheng, Tianheng and Wang, Xinjie and Lin, Tianwei and Su, Zhizhong and Liu, Wenyu and Wang, Xinggang},
  booktitle = {Proceedings of the IEEE/CVF Conference on Computer Vision and Pattern Recognition},
  pages     = {11960--11970},
  year      = {2025},
  doi       = {10.1109/CVPR52734.2025.01117},
  url       = {https://openaccess.thecvf.com/content/CVPR2025/html/Jiang_GaussTR_Foundation_Model-Aligned_Gaussian_Transformer_for_Self-Supervised_3D_Spatial_Understanding_CVPR_2025_paper.html}
}

@INPROCEEDINGS{cheng2022mask2former,
  author={Cheng, Bowen and Misra, Ishan and Schwing, Alexander G. and Kirillov, Alexander and Girdhar, Rohit},
  booktitle={2022 IEEE/CVF Conference on Computer Vision and Pattern Recognition (CVPR)}, 
  title={Masked-attention Mask Transformer for Universal Image Segmentation}, 
  year={2022},
  volume={},
  number={},
  pages={1280-1289},
  doi={10.1109/CVPR52688.2022.00135}}

@inproceedings{tan2019efficientnet,
  title     = {{EfficientNet}: Rethinking Model Scaling for Convolutional Neural Networks},
  author    = {Tan, Mingxing and Le, Quoc V.},
  booktitle = {Proceedings of the 36th International Conference on Machine Learning},
  series    = {Proceedings of Machine Learning Research},
  volume    = {97},
  pages     = {6105--6114},
  year      = {2019},
  url       = {https://proceedings.mlr.press/v97/tan19a.html}
}

@inproceedings{barsellotti2025talking,
  title     = {Talking to {DINO}: Bridging Self-Supervised Vision Backbones with Language for Open-Vocabulary Segmentation},
  author    = {Barsellotti, Luca and Bianchi, Lorenzo and Messina, Nicola and Carrara, Fabio and Cornia, Marcella and Baraldi, Lorenzo and Falchi, Fabrizio and Cucchiara, Rita},
  booktitle = {Proceedings of the IEEE/CVF International Conference on Computer Vision},
  pages     = {22025--22035},
  year      = {2025},
  doi       = {10.1109/ICCV51701.2025.02045},
  url       = {https://openaccess.thecvf.com/content/ICCV2025/html/Barsellotti_Talking_to_DINO_Bridging_Self-Supervised_Vision_Backbones_with_Language_for_ICCV_2025_paper.html}
}

@inproceedings{yin2023metric3d,
  title     = {{Metric3D}: Towards Zero-shot Metric {3D} Prediction from A Single Image},
  author    = {Yin, Wei and Zhang, Chi and Chen, Hao and Cai, Zhipeng and Yu, Gang and Wang, Kaixuan and Chen, Xiaozhi and Shen, Chunhua},
  booktitle = {Proceedings of the IEEE/CVF International Conference on Computer Vision},
  pages     = {9043--9053},
  year      = {2023},
  doi       = {10.1109/ICCV51070.2023.00830},
  url       = {https://openaccess.thecvf.com/content/ICCV2023/html/Yin_Metric3D_Towards_Zero-shot_Metric_3D_Prediction_from_A_Single_Image_ICCV_2023_paper.html}
}

@article{oquab2024dinov2,
  title   = {{DINOv2}: Learning Robust Visual Features without Supervision},
  author  = {Oquab, Maxime and Darcet, Timoth{\'e}e and Moutakanni, Th{\'e}o and Vo, Huy V. and Szafraniec, Marc and Khalidov, Vasil and Fernandez, Pierre and Haziza, Daniel and Massa, Francisco and El-Nouby, Alaaeldin and Assran, Mido and Ballas, Nicolas and Galuba, Wojciech and Howes, Russell and Huang, Po-Yao and Li, Shang-Wen and Misra, Ishan and Rabbat, Michael and Sharma, Vasu and Synnaeve, Gabriel and Xu, Hu and J{\'e}gou, Herv{\'e} and Mairal, Julien and Labatut, Patrick and Joulin, Armand and Bojanowski, Piotr},
  journal = {Transactions on Machine Learning Research},
  year    = {2024},
  url     = {https://openreview.net/forum?id=a68SUt6zFt}
}

@inproceedings{satopaa2011kneedle,
  author    = {Satopaa, Ville and Albrecht, Jeannie and Irwin, David and Raghavan, Barath},
  booktitle = {2011 31st International Conference on Distributed Computing Systems Workshops},
  title     = {Finding a ``{Kneedle}'' in a Haystack: Detecting Knee Points in System Behavior},
  year      = {2011},
  pages     = {166--171},
  doi       = {10.1109/ICDCSW.2011.20}
}

@article{mclachlan2019mixture,
  title     = {Finite Mixture Models},
  author    = {McLachlan, Geoffrey J. and Lee, Sharon X. and Rathnayake, Suren I.},
  journal   = {Annual Review of Statistics and Its Application},
  volume    = {6},
  number    = {1},
  pages     = {355--378},
  year      = {2019},
  publisher = {Annual Reviews}
}

@inproceedings{tang2024sparseocc,
  author    = {Tang, Pin and Wang, Zhongdao and Wang, Guoqing and Zheng, Jilai and Ren, Xiangxuan and Feng, Bailan and Ma, Chao},
  booktitle = {2024 IEEE/CVF Conference on Computer Vision and Pattern Recognition (CVPR)},
  title     = {{SparseOcc}: Rethinking Sparse Latent Representation for Vision-Based Semantic Occupancy Prediction},
  year      = {2024},
  pages     = {15035--15044},
  doi       = {10.1109/CVPR52733.2024.01424}
}

@inproceedings{wang2024opus,
  author    = {Wang, Jiabao and Liu, Zhaojiang and Meng, Qiang and Yan, Liujiang and Wang, Ke and Yang, Jie and Liu, Wei and Hou, Qibin and Cheng, Ming-Ming},
  booktitle = {Advances in Neural Information Processing Systems},
  doi       = {10.52202/079017-3809},
  editor    = {A. Globerson and L. Mackey and D. Belgrave and A. Fan and U. Paquet and J. Tomczak and C. Zhang},
  pages     = {119861--119885},
  publisher = {Curran Associates, Inc.},
  title     = {{OPUS}: Occupancy Prediction Using a Sparse Set},
  url       = {https://proceedings.neurips.cc/paper_files/paper/2024/file/d8ca28a32c05cd3b9b0940e43720f31b-Paper-Conference.pdf},
  volume    = {37},
  year      = {2024}
}
\bibliographystyle{iclr2027_conference}

\appendix

\setcounter{table}{0}\renewcommand{\thetable}{A\arabic{table}}
\setcounter{figure}{0}\renewcommand{\thefigure}{A\arabic{figure}}
\section{Experimental protocols and supporting evidence}\label{sec:appendix-evidence}

Throughout the appendices, GW, GF1, and GF3D abbreviate GaussianWorld, GaussianFormer-1, and GaussianFormer3D. Metric values are percentages unless a fraction is explicitly reported on the $[0,1]$ scale; changes are in percentage points. \emph{Decision drift} is the fraction of voxels occupied by the full model whose occupied/semantic decision changes after compilation. \emph{Mass recall} is the fraction of full-model Mean-FLIP mass assigned to rows retained by a router. CVaR denotes the conditional mean in the upper tail of a row's framewise scores; the exploratory \emph{composite} in Appendix~\ref{sec:temporal} combines mean score, CVaR, and activation frequency. A \emph{selective SSM} is a causal state-space router whose input-dependent transitions predict the rows retained at readout time.

\subsection{Evaluation protocols and checkpoint fidelity}\label{sec:protocols}

We use each repository's released evaluation code, with the protocol deviations listed below. Inspecting the checkpoints and implementations identifies the following eligible tensors:

\begin{itemize}[leftmargin=*]
\item \textbf{GaussianWorld:} all 25,600 persistent rows are eligible.
\item \textbf{GaussianFormer-1:} all 25,600 anchor/feature rows are eligible.
\item \textbf{OccFormer~\citep{zhang2023occformer} (non-Gaussian transfer, SemanticKITTI~\citep{behley2019semantickitti}):} all 100 persistent Mask2Former \citep{cheng2022mask2former} queries are eligible; evaluated with its official semantic scene completion mIoU (SSC mIoU, 19 classes) and scene completion IoU (SC IoU) on the SemanticKITTI validation sequence 08 (815 frames, 256$\times$256$\times$32 voxels).
\item \textbf{GaussianFormer3D~\citep{zhao2025gaussianformer3d} (LiDAR+camera Gaussian bank, nuScenes-SurroundOcc):} all 25,600 persistent Gaussian anchors are eligible; evaluated with the SurroundOcc 17-class mIoU and geometric IoU on the 6,019 nuScenes validation keyframes. Protocol deviation: the released recipe aggregates 10 LiDAR sweeps and consumes a ground-truth depth map derived from them; the nuScenes sweep files are not available in our environment, so every arm (released, compiled, controls, recovered) uses keyframe LiDAR only and a depth map projected from that keyframe. Absolute numbers are therefore below the README value and are compared only across our own arms.
\item \textbf{GaussianFlowOcc (camera-only Gaussian bank with temporal flow, Occ3D-nuScenes~\citep{tian2023occ3d}):} all 10,000 persistent slots (initial means and query features) are eligible; evaluated with the repository's Occ3D mIoU (15 classes, excluding \texttt{\detokenize{others}} and \texttt{\detokenize{other flat}}; camera-visible mask) and occupied IoU on the 6,019 validation frames. Protocol deviation: no checkpoint is released, so we trained the repository's own 3D-supervised path and calibrated the full model's occupancy threshold ($\tau=0.2$); Appendix~\ref{sec:gflow-results} gives the complete recipe. Absolute numbers are therefore not comparable to the GaussianFlowOcc paper and are compared only across our own arms.
\end{itemize}

Absolute accuracy and latency are compared only within an architecture. Unless a row explicitly states otherwise, reported latency uses PyTorch fp32, batch size 1, on one NVIDIA L4. TensorRT, mixed precision, and deployment-specific kernels may change both absolute latency and relative savings. Every Table~\ref{tab:main-summary} peak-memory value uses the same primary metric: fresh-process \texttt{\detokenize{torch.cuda.max_memory_allocated()}} after model warm-up and a peak-statistics reset, with full and compiled shapes resident separately. Allocator-reserved and independently sampled NVML process peaks are retained only as diagnostics and are never mixed into the headline column.

\textbf{Same-$K$ proxy definitions.} Each control keeps the adaptive arm's survivor count and changes only the ranking. Random uses seed 0. Fire/activity is the calibration-mean activity under the model's native gate (Gaussian opacity above the scorer's 0.1 fire threshold, or query-mask probability above 0.5). Opacity is median predicted opacity over calibration frames. Magnitude/contribution is the implementation's non-counterfactual readout-mass score; query norm is the joint $\ell_2$ norm of the aligned learned query tables. Taylor ranks each row by the calibration mean of $\sum_{p\in i}\lvert w_p\,\partial\mathcal L/\partial w_p\rvert$ over its aligned parameters and uses training labels; the other criteria are label-free. All data-dependent controls use training calibration inputs, and every arm is compiled and evaluated with the same strict-load and official-validation protocol as Exact Mean-FLIP.

Every model except GaussianFlowOcc starts from a public checkpoint; GaussianFlowOcc uses the supervised reproduction described in Appendix~\ref{sec:gflow-results}. The checkpoint-provenance table reports, per architecture and under one evaluator, the released checkpoint, the source checkpoint that the pipeline compiles, and the matched full control. The first two anchor our evaluators to published numbers and separate the effect of the recovery epoch from the effect of compilation; they are references, not controls, because they receive no recovery epoch.

\begin{table}[t]
\caption{Checkpoint and evaluator provenance}\label{tab:released}
\centering
\footnotesize
\setlength{\tabcolsep}{3pt}
\fitwidth{%
\begin{tabular}{>{\raggedright\arraybackslash}p{0.20\linewidth}>{\raggedright\arraybackslash}p{0.17\linewidth}>{\raggedright\arraybackslash}p{0.16\linewidth}>{\raggedright\arraybackslash}p{0.18\linewidth}>{\raggedright\arraybackslash}p{0.22\linewidth}}
\toprule
\textbf{Model} & \textbf{Released checkpoint (README mIoU)} & \textbf{Released, our evaluator, 0 ep (mIoU / geo IoU)} & \textbf{Source checkpoint compiled by the pipeline, 0 ep} & \textbf{Matched full, 1 recovery epoch} \\
\midrule
GaussianWorld & released streaming checkpoint (22.13) & 21.759 / 32.804 & Opacity-calibrated source checkpoint: 21.902 / 33.140 & 21.874$\pm$0.025 / 33.097$\pm$0.053 \\
GaussianFormer-1 & \path{NonEmpty} 25,600 solid (19.31) & 17.571 / 28.779 & Same as released & 19.291$\pm$0.002 / 30.050$\pm$0.002 \\
OccFormer (SemanticKITTI) & \path{occformer_kitti.pth} (13.50 SSC mIoU) & 13.50 / 36.42 (SSC mIoU / SC IoU) & Same as released & 12.99 / 36.22 (paired evaluation; one epoch, lr 5e-5, official 8-GPU batch) \\
GaussianFormer3D (nuScenes-SurroundOcc, keyframe LiDAR only) & \path{surroundocc_release.pth} & 18.40 / 30.06 (keyframe-only inputs; README value uses 10 LiDAR sweeps) & Same as released & 21.152 / 35.496 (one epoch, lr 4e-5, image backbone frozen; Appendix~\ref{sec:gf3d-results} explains the +2.75 mIoU adaptation gain) \\
GaussianFlowOcc (Occ3D-nuScenes~\citep{tian2023occ3d}, our 3D-supervised reproduction, $\tau$ = 0.2) & none released; ours trained for 18 epochs (weak-recipe paper numbers not comparable) & 24.65 / 54.86 (mIoU / occupied IoU; 20.79 / 38.54 at the weak-recipe default $\tau$ = 0.05) & Same as trained & 24.70 / 54.75 \\
\bottomrule
\end{tabular}%
}
\end{table}

Our evaluators track the released baselines, while the large recovery-only shifts on GF1 and GF3D show why every pruning result must be compared with an identically trained full control.

The released GaussianWorld checkpoint scores 21.759 under the streaming protocol (38-frame windows over 150 validation scenes, 5,700 frames) against a README value of 22.13, so the evaluator is faithful. The released GaussianFormer-1 weights score 1.74 points below their README value in the same evaluator and recover to 19.291 after one matched unpruned epoch on our data pipeline; we attribute the gap to pipeline adaptation and report it rather than correct for it. Every compiled result is therefore compared with a control that receives the identical recovery epoch: for GaussianFormer-1 the epoch alone moves semantic mIoU by +1.72 points, whereas compilation costs 0.17 point. Peak memory is measured in a fresh process. Every adaptive result records the Otsu threshold, separation $\eta$, survivor indices, strict-load certificate, and official validation metrics. Appendix~\ref{sec:gw-results} distinguishes the lifecycle and static GaussianWorld protocols.

\subsection{Architecture-specific protocols, results, and interpretation}

\subsubsection{GaussianWorld}\label{sec:gw-results}

\textbf{Protocol.} The static reference scores isolated frames and is reported separately from the primary streaming lifecycle protocol.

\begin{table}[t]
\caption{GaussianWorld static-frame deployment reference}\label{tab:gw-static}
\centering
\footnotesize
\setlength{\tabcolsep}{3pt}
\fitwidth{%
\begin{tabular}{lrr}
\toprule
\textbf{Stage / metric} & \textbf{Matched full $K=25{,}600$} & \textbf{Adaptive $K=16{,}461$} \\
\midrule
Otsu separation $\eta$ & N/A & 0.6361 \\
Automatically removed & 0 & 9,139 (35.699\%) \\
Image backbone & 514.38 ms & 513.65 ms (-0.1\%) \\
Neck & 7.81 ms & 7.83 ms (+0.2\%) \\
Encoder & 170.35 ms & 120.74 ms (-29.1\%) \\
-- camera sampling (DFA) & 111.67 ms & 74.91 ms (-32.9\%) \\
Head incl. aggregation & 146.94 ms & 119.40 ms (-18.7\%) \\
-- local aggregation & 68.45 ms & 42.86 ms (-37.4\%) \\
\textbf{Gaussian-side subtotal} & \textbf{317.29 ms} & \textbf{240.14 ms (-24.3\%)} \\
\textbf{Whole model} & \textbf{839.80 ms} & \textbf{761.92 ms (-9.3\%)} \\
Peak GPU memory & 5.792 GiB & 5.791 GiB \\
Semantic mIoU / Geometry IoU, compiled, 0 recovery epochs & 21.902 / 33.140 (calibrated source) & 21.490 / 32.566 (-0.412 / -0.574) \\
Semantic mIoU & 21.874$\pm$0.025 & \textbf{21.649} \\
Geometry IoU & 33.097$\pm$0.053 & \textbf{32.827} \\
\bottomrule
\end{tabular}%
}
\end{table}

\textbf{Quantitative result.} Static Mean-FLIP removes 35.7\% of the Gaussians and cuts Gaussian-side latency by 24.3\% and end-to-end latency by 9.3\%, at a cost of 0.225 mIoU and 0.270 geometry IoU against the recovered full control. Before recovery the gap is 0.41 / 0.57, so the matched epoch closes 0.16 / 0.26 of it. The adaptive latency is the mean of two 100-round profiles (756.61 and 767.22 ms, a 1.39\% difference). This is a deployment reference, not the primary lifecycle-aware result.

\begin{table}[t]
\caption{GaussianWorld lifecycle-integrated compilation}\label{tab:gw-lifecycle}
\centering
\footnotesize
\setlength{\tabcolsep}{3pt}
\fitwidth{%
\begin{tabular}{lrr}
\toprule
\textbf{Stage / metric} & \textbf{Matched full $K=25{,}600$} & \textbf{Lifecycle-adaptive $K=22{,}427$} \\
\midrule
Otsu separation $\eta$ & N/A & \textbf{0.8406} \\
Automatically removed & 0 & \textbf{3,173 (12.395\%)} \\
Image backbone & 529.52 ms & 529.06 ms (-0.1\%) \\
Neck & 7.86 ms & 7.87 ms (+0.2\%) \\
Encoder & 171.29 ms & 154.98 ms (-9.5\%) \\
-- camera sampling (DFA) & 112.90 ms & 100.58 ms (-10.9\%) \\
Head incl. aggregation & 99.61 ms & 96.96 ms (-2.7\%) \\
-- local aggregation & 89.64 ms & 87.80 ms (-2.1\%) \\
\textbf{Gaussian-side subtotal} & \textbf{270.90 ms} & \textbf{251.94 ms (-7.0\%)} \\
\textbf{Whole model} & \textbf{808.54 ms} & \textbf{789.13 ms (-2.4\%)} \\
Peak GPU memory & 5.854 GiB & 5.782 GiB \\
Semantic mIoU / Geometry IoU, compiled, 0 recovery epochs & 21.902 / 33.140 (calibrated source) & \textbf{21.723 / 32.850 (-0.179 / -0.290)} \\
Semantic mIoU, matched one epoch & 21.874$\pm$0.025 & \textbf{21.792 (-0.082)} \\
Geometry IoU, matched one epoch & 33.097$\pm$0.053 & \textbf{32.978 (-0.119)} \\
Recovery gain of compiled model & --- & \textbf{+0.069 / +0.128} \\
Evaluation coverage & 5,700 streaming frames & 5,700 streaming frames \\
\bottomrule
\end{tabular}%
}
\end{table}

\textbf{Interpretation.} The lifecycle run follows each initial learned row through streaming state updates and refill boundaries, applies the same guarded Otsu rule without a target survivor count, and locks both \texttt{\detokenize{lifter.num_anchor}} and \texttt{\detokenize{future_decoder.num_anchor}} to 22,427. Its automatic cut is substantially more conservative than the static-frame cut above and reduces the post-recovery gap to 0.082 mIoU / 0.119 geometry IoU. On the local NVIDIA L4-24Q, two independent 100-round round-robin profiles give whole-model means of 809.10 / 807.98 ms for the full model and 788.83 / 789.42 ms for the lifecycle-compiled model. The cross-run differences are 0.14\% and 0.07\%, respectively, passing the prespecified 1\% guard; each run also passes the 2\% mean-versus-median guard. These latencies are paired within the streaming configuration of Table~\ref{tab:gw-lifecycle}; the static reference's absolute latencies belong to its separate configuration and $K=16{,}461$ budget and are not comparable.

\subsubsection{GaussianFormer-1}\label{sec:gf1-results}

\textbf{Protocol.} The scorer uses 500 training-calibration frames; compilation is strict-loaded, and each fixed-checkpoint evaluation covers all 753 distributed validation batches. The frozen $K=17{,}365$ mask is shared across recovery seeds.

\begin{table}[t]
\caption{GaussianFormer-1 compilation, efficiency, and matched recovery}\label{tab:gf1}
\centering
\footnotesize
\setlength{\tabcolsep}{3pt}
\fitwidth{%
\begin{tabular}{lrr}
\toprule
\textbf{Stage / metric} & \textbf{Matched full $K=25{,}600$} & \textbf{Adaptive lifecycle Mean-FLIP + Otsu} \\
\midrule
Otsu separation $\eta$ & N/A & \textbf{0.6593} \\
Automatically selected $K$ & 25,600 & \textbf{17,365} \\
Automatically removed & 0 & \textbf{8,235 (32.168\%)} \\
Image backbone & 535.66 ms & 533.67 ms (-0.4\%) \\
Neck & 7.75 ms & 7.75 ms (+0.0\%) \\
Encoder & 123.53 ms & 89.29 ms (-27.7\%) \\
-- camera sampling (DFA) & 86.52 ms & 61.06 ms (-29.4\%) \\
Head incl. aggregation & 31.45 ms & 29.54 ms (-6.1\%) \\
-- local aggregation & 21.31 ms & 19.91 ms (-6.6\%) \\
\textbf{Gaussian-side subtotal} & \textbf{154.98 ms} & \textbf{118.83 ms (-23.3\%)} \\
\textbf{Whole model} & \textbf{698.61 ms} & \textbf{660.44 ms (-5.5\%)} \\
Peak GPU memory & 2.031 GiB & 2.048 GiB \\
A100 whole-model diagnostic & 230.1 ms & 216.6 ms (-5.9\%) \\
A100 peak GPU memory diagnostic & 2.03 GiB & 2.05 GiB \\
Semantic mIoU / Geometry IoU, compiled, 0 recovery epochs & 17.571 / 28.779 (released weights) & 17.414 / 28.344 (-0.157 / -0.435) \\
Semantic mIoU, three-seed matched recovery & 19.291$\pm$0.002 & \textbf{19.120$\pm$0.003 (-0.171$\pm$0.002)} \\
Geometry IoU, three-seed matched recovery & 30.050$\pm$0.002 & \textbf{29.804$\pm$0.001 (-0.246$\pm$0.001)} \\
\bottomrule
\end{tabular}%
}
\end{table}

\textbf{Quantitative result.} On GaussianFormer-1, removing 32.2\% of the bank cuts Gaussian-side latency by 23.3\% and whole-model latency by 5.5\%. Across matched recovery seeds 0, 1, and 2, the difference is $-0.171\pm0.002$ mIoU and $-0.246\pm0.001$ geometry IoU; no checkpoint is selected by validation performance.

Because every GF1 row is alive on every scored frame, the lifecycle integral of Section~\ref{sec:lifecycle} equals the saved Mean-FLIP score up to float32 accumulation error (maximum absolute difference $1.94\times10^{-6}$); its Otsu mask and measured recovery results are therefore unchanged. Every seed uses the same frozen mask and covers all 753 validation batches. No validation checkpoint was selected.

\textbf{Interpretation.} The same run also contains a fixed-$K=17{,}296$ exact arm and matched opacity and random controls at that budget; before recovery they score 17.406 / 28.318 (exact, fixed $K$), 17.384 / 27.545 (opacity ranking) and 9.142 / 11.648 (random ranking), and after the matched epoch 19.108 / 29.698, 19.105 / 29.698 and 18.504 / 28.711. Compilation alone therefore costs the exact ranking 0.16 mIoU / 0.44 IoU at a 32\% cut, whereas a random cut of the same size collapses the model before recovery and remains 0.6 mIoU below the exact ranking after it.

The L4 profile uses one NVIDIA L4-24Q at batch size 1 with both configurations resident and interleaved round-robin: 10 warm-up and 100 measured rounds in each of two runs. Whole-model means were 697.31 / 699.90 ms (full) and 658.63 / 662.26 ms (compiled), with inter-run variation of 0.37\% and 0.55\% and mean-versus-median gaps no larger than 0.70\%. The full arm loads the released checkpoint; the compiled arm strictly loads the saved $K=17{,}365$ structure. Peak memory is measured in a fresh process. The A100 measurements are diagnostic and are not combined with the L4 protocol.

\subsubsection{OccFormer}

\textbf{Protocol.} The 100 persistent Mask2Former queries are scored on 200 training frames (20 per sequence, seed 0) at the full $256\times256\times32$ decision volume; labels are not used. Compiled checkpoints are strict-loaded and evaluated on all 815 frames of SemanticKITTI sequence 08.

\begin{table}[t]
\caption{OccFormer query compilation, efficiency, and matched recovery}\label{tab:occformer}
\centering
\footnotesize
\setlength{\tabcolsep}{3pt}
\fitwidth{%
\begin{tabular}{>{\raggedright\arraybackslash}p{0.33\linewidth}>{\raggedleft\arraybackslash}p{0.15\linewidth}>{\raggedleft\arraybackslash}p{0.21\linewidth}>{\raggedleft\arraybackslash}p{0.24\linewidth}}
\toprule
\textbf{Stage / metric} & \textbf{Matched full $K=100$} & \textbf{Fixed-$K$ reference, $K=60$ (same flip ranking)} & \textbf{Adaptive Exact Mean-FLIP + Otsu} \\
\midrule
Eligible population & 100 persistent Mask2Former queries & Same population & Same population \\
Otsu separation $\eta$ & N/A & N/A (fixed budget) & \textbf{0.795} (69 queries change $\leq$1 voxel per frame; 3 exactly zero) \\
Automatically selected $K$ & 100 & 60 (fixed) & \textbf{21} \\
Automatically removed & 0 & 40 (40.0\%, fixed) & \textbf{79 (79.0\%)} \\
Image backbone (EfficientNet-B7) & 61.43 ms & 61.56 ms (+0.2\%) & 61.44 ms (+0.0\%) \\
Image neck (SECOND FPN) & 0.45 ms & 0.45 ms & 0.45 ms \\
LSS voxel lift & 18.45 ms & 18.44 ms (-0.1\%) & 18.40 ms (-0.3\%) \\
3D occupancy encoder, conv backbone & 260.69 ms & 261.18 ms (+0.2\%) & 260.68 ms (-0.0\%) \\
3D occupancy encoder, deformable-attention neck & 217.72 ms & 217.54 ms (-0.1\%) & 217.86 ms (+0.1\%) \\
Mask2Former head (queries + mask readout) & 53.93 ms & 41.58 ms (-22.9\%) & 29.75 ms (-44.8\%) \\
-- nine query decoder layers & 22.87 ms & 18.13 ms (-20.7\%) & 13.60 ms (-40.5\%) \\
-- per-query mask logits inside the head forward (mask embedding $\times$ mask feature) & 30.62 ms & 22.96 ms (-25.0\%) & 15.85 ms (-48.2\%) \\
Readout outside the head module: trilinear upsampling of every per-query mask to 256$\times$256$\times$32, class-weighted sum, argmax (residual = whole minus hooked modules) & 26.70 ms & 20.72 ms (-22.4\%) & 14.62 ms (-45.2\%) \\
\textbf{Query-dependent subtotal (head + readout)} & \textbf{80.63 ms} & \textbf{62.30 ms (-22.7\%)} & \textbf{44.37 ms (-45.0\%)} \\
\textbf{Whole model} & \textbf{639.37 ms} & \textbf{621.47 ms (-2.8\%)} & \textbf{603.20 ms (-5.7\%)} \\
Peak GPU memory & 3.704 GiB & 3.089 GiB (-16.6\%) & 3.089 GiB (-16.6\%) \\
SSC mIoU / SC IoU, released checkpoint, 0 recovery epochs & 13.50 / 36.42 & \textbf{13.60 / 36.42 (+0.10)} (compiled, before recovery) & \textbf{13.34 / 36.42 (-0.16)} (compiled, before recovery) \\
Same-$K$ random control, 0 recovery epochs & N/A & 4.39 / 14.60 & 0.82 / 7.89 \\
Same-$K$ fire-rate control, 0 recovery epochs & N/A & 13.62 / 36.42 & 13.17 / 36.41 (bicycle, motorcycle, traffic-sign fall to 0.00) \\
SSC mIoU / SC IoU, paired matched epoch (lr 5e-5, 100-iteration warm-up, official 8-GPU SyncBN batch) & 12.99 / 36.22 & N/A & \textbf{13.02 / 36.30 (+0.03 / +0.08 vs full)} \\
\bottomrule
\end{tabular}%
}
\end{table}

\textbf{Quantitative result.} OccFormer's persistent queries are rankable: a conservative 40\% cut matches or exceeds the full control, while the automatic 79\% cut reduces query-dependent work by 45\% and whole-model latency by 5.7\% with a small, recovery-sensitive mIoU change.

\textbf{Interpretation.} Inspection of the released OccFormer \citep{zhang2023occformer} SemanticKITTI implementation shows a single-image EfficientNet-B7 \citep{tan2019efficientnet} backbone, a lift-splat-shoot voxel lift, a 3D occupancy encoder (3D convolutions followed by a 3D deformable-attention neck), and a Mask2Former head with 100 persistent learnable queries (\texttt{\detokenize{query_embed}} and \texttt{\detokenize{query_feat}}, the only per-query tables in the checkpoint). Its inference readout is additive over queries: $\text{score}[c,x]=\sum_q \mathrm{softmax}(\text{cls}_q)[c]\,\sigma(\text{mask}_q[x])$ with the label $\arg\max_c$ over 20 classes (class 0 is empty), so deleting a query is an exact subtraction from the cached scores. Sixty-nine of the 100 queries change at most one voxel decision per frame and carry a no-object probability above 0.99, three change none, and one query alone flips 83\% of the voxels because it carries the empty class.

The Otsu valley on log10 scores falls at $K=21$ (split-half thresholds 0.56 / 0.71 in log10 units), so the adaptive rule removes 79 of the 100 queries without a target count; the fixed $K=60$ column removes only the 40 lowest-ranked queries as a reference. At $K=60$ the compiled model is not worse than the released checkpoint before any recovery (13.60 versus 13.50 SSC mIoU, SC IoU unchanged), whereas the same-$K$ random control collapses to 4.39, so the ranking carries signal. At the adaptive budget the compiled model loses 0.16 point with SC IoU unchanged; the fire-rate control at the same $K=21$ loses 0.33 point and erases three rare classes (bicycle, motorcycle, traffic-sign) that the exact flip score keeps, and the random control collapses to 0.82.

The paired evaluation gives every arm the same one-epoch recovery under the official training batch (8 GPUs $\times$ 1 sample, synchronized BatchNorm), half base learning rate (5e-5), and 100-iteration warm-up. It evaluates all 815 frames: the matched full reaches 12.99 / 36.22 and adaptive $K=21$ reaches 13.02 / 36.30, a compiled-minus-full change of +0.03 / +0.08. Before recovery the compiled adaptive model is within 0.16 point of the released checkpoint, so for OccFormer compilation itself needs no recovery. Our evaluator reproduces the README value of the released checkpoint exactly (13.50 SSC mIoU; the released training log peaks at 13.50 in epoch 21).

\emph{L4 latency.} One NVIDIA L4-24Q, PyTorch 1.10.1 + CUDA 11.3, fp32, batch size 1, one 384$\times$1280 left image (validation frame 08/000500), the three query budgets resident and timed round-robin (10 warm-up + 100 measured rounds, mean of two runs; whole-model runs 639.26 / 639.48 ms, 621.36 / 621.58 ms and 603.11 / 603.29 ms, inter-run deviation below 0.05\%; the head breakdown rows come from a third identical run whose head totals agree within 0.7 ms). Compiled query banks are the strict-loaded sliced checkpoints; latency does not depend on the weight values. The query-dependent work is the head forward (nine decoder layers and the per-query mask logits) plus the readout that OccFormer runs outside the head module: every per-query mask is upsampled to the 256$\times$256$\times$32 decision grid, weighted by its class probabilities, summed and arg-maxed. The hooks do not cover that readout, so we report it as the residual between the whole-model time and the sum of hooked modules; it shrinks from 26.7 to 14.6 ms as the query count falls, consistent with that attribution. Head and readout together hold 12.6\% of the per-frame time and fall by 23\% at $K=60$ and 45\% at $K=21$; Amdahl's law then predicts a 5.7\% whole-model saving at $K=21$, matching the measured 5.66\%. The 3D occupancy encoder (conv backbone 40.8\%, deformable-attention neck 34.1\%) and the image backbone (9.6\%) do not depend on the query count, so the whole-model saving is 2.8\% and 5.7\%, respectively. Peak memory drops from 3.704 GiB at $K=100$ to 3.089 GiB at $K=60$, then remains at 3.089 GiB at $K=21$. Reducing query-mask storage causes the initial decrease, after which fixed activations and CUDA allocator granularity determine the measured peak. The transfer claim is an accuracy and selection claim: the exact flip score identifies the 69 near-idle queries (three exactly zero) without a target count, and at a 79\% cut it preserves rare classes that the fire-rate proxy discards.

\subsubsection{GaussianFormer3D}\label{sec:gf3d-results}

\textbf{Protocol.} Every arm uses keyframe LiDAR only, as specified in Appendix~\ref{sec:protocols}. Scores are accumulated on 500 training keyframes at stride 12, and all reported comparisons use the complete 6,019-frame validation set.

\begin{table}[t]
\caption{GaussianFormer3D compilation under the keyframe-LiDAR protocol}\label{tab:gf3d}
\centering
\footnotesize
\setlength{\tabcolsep}{3pt}
\fitwidth{%
\begin{tabular}{lrr}
\toprule
\textbf{Stage / metric} & \textbf{Matched full $K=25{,}600$} & \textbf{Adaptive Exact Mean-FLIP + guarded Otsu $K=22{,}535$} \\
\midrule
Otsu separation $\eta$ & N/A & \textbf{0.569} (gate 0.5 passed; log10 threshold -1.060, equivalent to 0.0872 flips/frame) \\
Automatically removed & 0 & \textbf{3,065 (12.0\%)} \\
Gaussian encoder & 382.8 ms & \textbf{332.2 ms (-13.2\%)} \\
-- DeformableFeatureAggregation3D & 328.3 ms & \textbf{283.6 ms (-13.6\%)} \\
Head incl. local aggregation readout & 151.2 ms & 150.0 ms (-0.8\%) \\
-- local aggregation & 73.4 ms & 70.7 ms (-3.7\%) \\
Slot-independent modules & 707--710 ms & Unchanged within noise \\
\textbf{Whole model, L4, batch 1} & \textbf{1230.0 ms} & \textbf{1172.6 ms (-4.7\%)} \\
Whole-model median & 1160.4 ms & 1107.3 ms (-4.6\%) \\
Peak GPU memory & 3.810 GiB & \textbf{3.492 GiB (-8.3\%)} \\
Semantic mIoU / Geometry IoU, 0 recovery epochs & 18.40 / 30.06 & \textbf{18.36 / 29.96 (-0.04 / -0.10)} \\
Semantic mIoU / Geometry IoU, matched one epoch & 21.152 / 35.496 & \textbf{21.128 / 35.380 (-0.024 / -0.116)} \\
\bottomrule
\end{tabular}%
}
\end{table}

\textbf{Quantitative result.} On GaussianFormer3D, the automatically selected 12\% cut has a single-seed difference of -0.024 mIoU / -0.116 geometry IoU while reducing whole-model latency by 4.7\% and peak memory by 8.3\%.

\textbf{Interpretation.} The same-$K$ controls for GaussianFormer3D (exact, fire-rate, random at $K=22{,}535$) are reported in Table~\ref{tab:samek-full}. Every compiled arm there retains the same slot count, so its latency and memory structure is identical and the comparison isolates ranking quality.

GaussianFormer3D keeps the GaussianFormer-1 anchor bank (28 anchor parameters and a 128-d instance feature per slot) and readout (linear local aggregation), and adds a LiDAR branch (voxelization, sparse convolution) and a depth head whose ground-truth depth input is projected from LiDAR. The bank therefore satisfies both structural conditions, and the exact removal identity is the one verified for GaussianFormer-1. The LiDAR and image branches do not depend on the slot count, so the whole-model saving is again bounded by the encoder and readout share of latency. Flip scores are accumulated on 500 training keyframes (stride 12). The analytic reconstruction mirrors the aggregation kernel's per-Gaussian box (radius $\lceil 3\sigma_{\max}/0.5\,\mathrm{m}\rceil$ voxels, up to 11 for this model's 1.8 m scale cap, versus 4 for GaussianFormer-1) and is checked against brute-force delete-and-recompute with the official CUDA aggregator on the highest-contribution slots: maximum absolute logit difference $1.2\times10^{-5}$, zero voxels outside tolerance, argmax identical. The exact flip score correlates with the ground-truth hit rate (Spearman 0.86) and less with fire rate (0.77) and median opacity (0.72), the same pattern as GaussianFormer-1. The valley removes only 12\% of the bank: with a LiDAR branch feeding the anchors, far fewer slots are decision-idle than in the camera-only GaussianFormer banks. At that budget the compiled model loses 0.04 mIoU / 0.10 IoU, the same-$K$ fire-rate control 0.10 / 0.20, and the same-$K$ random control 0.73 / 1.62. All arms share the keyframe-only inputs described in Appendix~\ref{sec:protocols}.

Latency follows the same Amdahl decomposition as the other models: removing 12\% of the slots cuts the slot-proportional deformable aggregation by 13.6\% and the Gaussian encoder by 13.2\%, while the image backbone (54\% of the 1.23 s whole-model time on the L4 at 6$\times$928$\times$1600), the LiDAR voxelization and sparse convolution, and the depth head do not move, so the whole model gains 4.7\% (mean) / 4.6\% (median) and peak memory falls 8.3\%. Absolute per-module times on the shared L4 workstation vary by up to 20\% between the two runs; the relative savings agree within 0.3 points across runs, which is why they are reported as ratios of interleaved round-robin means.

The strict matched full control trained for one epoch under the recovery recipe reaches 21.152 / 35.496, far above the released 18.40 / 30.06: the released weights were trained with ten aggregated LiDAR sweeps, and a single epoch on keyframe-only inputs adapts the LiDAR branch and depth head to the sparser point cloud. This makes the matched control indispensable for GaussianFormer3D --- the compiled bank after its own recovery epoch must be read against 21.152, not against the released number. In the strict full-versus-compiled run, the compiled bank reaches 21.128 / 35.380, a 0.024 mIoU / 0.116 IoU reduction.

\subsubsection{GaussianFlowOcc}\label{sec:gflow-results}

\textbf{Protocol.} GaussianFlowOcc is the sole exception to the public-checkpoint starting condition: no checkpoint is released. We therefore use the repository's 3D-supervised training path and compare only matched arms from that reproduction. The sequential evaluation advances temporal state through 8,971 consecutive forwards while scoring 300 frames. Guarded Otsu selects the primary $K=8{,}628$ at $\eta=0.856$; Kneedle and largest-gap instead share the distinct $K=8{,}725$ mask.

\begin{table}[t]
\caption{GaussianFlowOcc compilation of the supervised reproduction}\label{tab:gaussianflowocc}
\centering
\footnotesize
\setlength{\tabcolsep}{3pt}
\fitwidth{%
\begin{tabular}{>{\raggedright\arraybackslash}p{0.36\linewidth}>{\raggedleft\arraybackslash}p{0.17\linewidth}>{\raggedleft\arraybackslash}p{0.41\linewidth}}
\toprule
\textbf{Stage / metric} & \textbf{Full $10{,}000$} & \textbf{Adaptive $K$} \\
\midrule
Eligible population & 10,000 persistent slots (initial mean + query feature) & Same population \\
Otsu separation $\eta$ & N/A & \textbf{0.856} \\
Automatically selected $K$ & 10,000 & \textbf{8,628} \\
Automatically removed & 0 & \textbf{1,372 (13.72\%)} \\
Primary $K=8{,}628$ L4 whole-model / slot-dependent latency & 318.9 / 231.1 ms & \textbf{293.5 / 206.5 ms (-8.0\% / -10.7\%)} \\
Primary $K=8{,}628$ peak GPU memory (torch allocated) & 1.486 GiB & \textbf{1.382 GiB (-7.0\%)} \\
Semantic mIoU, 0 recovery epochs ($\tau$ = 0.2) & 24.65 & \textbf{24.75 (+0.10)} \\
Occupied IoU, 0 recovery epochs ($\tau$ = 0.2) & 54.86 & \textbf{55.43 (+0.57)} \\
Canonical same-$K$ heuristic proxy controls and decision drift & --- & unavailable (Taylor: Table~\ref{tab:samek-full}) \\
Same quantities at the weak-recipe default $\tau$ = 0.05 (mis-calibrated; $K=8{,}861$ there) & 20.79 / 38.54 & 20.91 / 38.86 (+0.12 / +0.32), drift 3.1\%; random 20.90 / 39.88 (drift 4.5\%); fire-rate 21.48 / 46.50 (drift 13.9\%) \\
Semantic mIoU / occupied IoU, after matched recovery & 24.70 / 54.75 & \textbf{24.71 / 54.78 (+0.01 / +0.03)} \\
\bottomrule
\end{tabular}%
}
\end{table}

\textbf{Quantitative result.} On our supervised GaussianFlowOcc at its calibrated threshold, sequential guarded Otsu removes 13.72\% and changes mIoU / occupied IoU by +0.10 / +0.57 before recovery. Matched recovery reaches 24.71 / 54.78 versus the matched full 24.70 / 54.75 (+0.01 / +0.03). On the L4 protocol, the primary compiled model reduces the slot-dependent subtotal by 10.7\%, whole-model latency by 8.0\%, and torch peak allocated memory by 7.0\%. Primary $K=8{,}628$ heuristic same-$K$ proxy results remain unavailable; complete matched-recovery per-class results are in Table~\ref{tab:perclass-occ3d}.

\textbf{Interpretation.} GaussianFlowOcc keeps a bank of 10,000 slots, each an initial mean and a query feature that the decoder refines every frame, and reads the refined Gaussians out by splatting opacity-weighted contributions into a 200$\times$200$\times$16 grid. The density is the clamped sum of contributions, the semantic logit of a voxel is the opacity-weighted mean of the slots' logits, and a voxel is occupied when its density exceeds a threshold. The mean's denominator is a positive scalar per voxel, so the class argmax equals the argmax of the numerator, which is linear in each slot's term. Removing a slot is therefore exact: subtract its density and its weighted logits before the clamp, then apply the two post-processing steps (density below 0.001 set to zero, roof rows removed) as in the model's own inference. We verified this on the trained checkpoint against the model's own readout with the slot physically removed: 80 of 80 probed slots on two frames give identical decisions, the scorer's readout matches the model's on every voxel, and the density error is $3\times10^{-7}$. The temporal module carries the previous frame's Gaussians forward, so flip scores are accumulated with the model run sequentially while scoring one frame in 30 to keep temporal state identical to the evaluation path. The evaluation contains 8,971 sequential forwards and 300 scored frames.

No checkpoint is released for this model and the weakly-supervised recipe starts from about 826 GB of generated depth and open-vocabulary pseudo labels, so we trained the repository's own 3D-supervised path (binary cross-entropy on the voxel density and cross-entropy on the voxel semantics against Occ3D labels, temporal supervision on the flowed frames, 18 epochs, 8 GPUs, batch 16, learning rate $10^{-4}$). The repository's evaluation threshold of 0.05 belongs to the weakly-supervised model; for the supervised density it marks 57\% of the grid occupied. We therefore calibrated the threshold of the full model on 25 whole training scenes disjoint from the calibration frames: mIoU 23.2 at 0.05, 26.2 at 0.1, 27.7 at 0.15, 28.2 at 0.2 (maximum), 28.1 at 0.25 and 27.5 at 0.3, with occupied IoU peaking at 0.3; the threshold that maximizes mIoU, $\tau = 0.2$, is used for every arm. The full model then scores 24.65 mIoU / 54.86 IoU on the validation frames instead of 20.79 / 38.54. The mis-calibrated row of Table~\ref{tab:gaussianflowocc} shows why this matters for a pruning comparison: at 0.05 every deletion removes false-positive occupancy, so the same-$K$ fire-rate control gains 8 IoU while changing 13.9\% of the full model's occupied decisions, and a random cut gains 1.3 IoU. Only the exact flip cut leaves the decisions comparatively stable (drift 3.1\%, +0.12 / +0.32).

The score itself behaves as on the other Gaussian banks: the readout is both-type (emptied share 0.52), the flip score correlates with the ground-truth hit rate (Spearman 0.73) and with splatted mass (0.80) and barely with fire rate (0.12), and the two halves of the calibration frames rank the slots almost identically (split-half Spearman 0.985). The valley falls at the guarded-Otsu boundary and removes 1,372 slots (13.72\%). The primary compiled model scores 24.75 / 55.43 before recovery and 24.71 / 54.78 after matched recovery, compared with 24.65 / 54.86 and 24.70 / 54.75 for the corresponding full checkpoints.

The primary efficiency profile strict-loads the exact $K=8{,}628$ structure and uses two fresh-process 100-forward latency runs. The two runs give slot-dependent reductions of 10.7\% and 10.6\% and whole-model reductions of 8.0\% and 7.9\%. Separate memory runs reproduce the 1.486 versus 1.382 GiB allocated peaks in the table. Detailed checkpoint and run provenance will be released with the code.

\subsubsection{GaussTR boundary case without stable row identity}\label{sec:gausstr}

\textbf{Protocol.} Compilation is applied only as a boundary test because reference points are resampled on every forward pass and checkpoint rows therefore lack persistent primitive identity. All arms use the released evaluator without recovery.

\begin{table}[t]
\caption{GaussTR boundary case without stable row identity}\label{tab:gausstr}
\centering
\footnotesize
\setlength{\tabcolsep}{3pt}
\fitwidth{%
\begin{tabular}{lrr}
\toprule
\textbf{Stage / metric} & \textbf{Released full $K=300$ per view} & \textbf{Exact Mean-FLIP + Otsu (compiled, no recovery)} \\
\midrule
Eligible population & 300 query embeddings reused over six views (1,800 generated Gaussians; reference points resampled every forward) & Same query rows \\
Otsu separation $\eta$ & N/A & 0.64 (gate passes) \\
Automatically selected $K$ & 300 & 158 \\
Automatically removed & 0 & 142 (47.3\%) \\
Image backbone (GaussTR \citep{jiang2025gausstr}: frozen DINOv2 \citep{oquab2024dinov2} ViT-B/14) & 530.36 ms & 530.00 ms (-0.1\%) \\
Neck & 69.72 ms & 69.77 ms (+0.1\%) \\
Query decoder & 18.54 ms & 16.41 ms (-11.5\%) \\
Gaussian head incl. voxelizer & 418.35 ms & 233.55 ms (-44.2\%) \\
-- voxelizer (Gaussian-to-voxel) & 408.10 ms & 223.61 ms (-45.2\%) \\
\textbf{Query-dependent subtotal (decoder + head)} & \textbf{436.89 ms} & \textbf{249.96 ms (-42.8\%)} \\
\textbf{Whole model} & \textbf{1049.18 ms} & \textbf{862.06 ms (-17.8\%)} \\
Peak GPU memory & 2.272 GiB & 1.867 GiB \\
Semantic mIoU / IoU, compiled only & 12.28 / 44.66 (README 12.27) & 11.47 / 40.32 (-0.81 / -4.34 points) \\
Random ranking, same $K=158$ & --- & 11.47 / 40.31 \\
Fire-rate ranking, same $K=158$ & --- & 11.46 / 40.31 \\
\bottomrule
\end{tabular}%
}
\end{table}

\textbf{Quantitative result.} GaussTR is the negative control: without stable row identity, Decision-FLIP ranks no better than random, so its 17.8\% speedup is the effect of a smaller bank, not of identified redundancy.

\textbf{Interpretation.} In the released GaussTR \citep{jiang2025gausstr} Talk2DINO \citep{barsellotti2025talking} variant, 300 query embeddings are reused for each of six views, but their reference points are drawn at random on every forward pass. A checkpoint row therefore does not generate the same primitive across frames. The per-query deletion is still exact (20 verified queries match brute-force deletion) and the score distribution passes the separation gate ($\eta=0.64$, $K=158$); but at the same $K$ the exact ranking, a random ranking, and fire rate are indistinguishable (11.47 / 11.47 / 11.46 mIoU), and accuracy degrades monotonically with the number of removed queries regardless of which are removed (fixed $K=200$: 11.83 / 42.45; fixed $K=150$: 11.38 / 39.78). This is the failure mode predicted by Section~\ref{sec:banks} when the removable unit has no persistent identity. The separation gate does not detect it, which is why eligibility is an architectural condition rather than a score statistic, and why the method makes no compilation claim on such models. Metrics are mIoU / IoU on the Occ3D-nuScenes~\citep{tian2023occ3d} validation split; no recovery epoch is applied to any arm.

\emph{L4 latency.} Same protocol as the other models (one L4-24Q, PyTorch 2.1.2, fp32, batch size 1, both budgets resident, round-robin, 10 warm-up + 100 measured rounds, mean of two runs; whole-model runs 1047.97 / 1050.39 ms and 860.59 / 863.53 ms, inter-run deviation 0.2\% and 0.3\%). The timed call is the model's own predict path on one calibration frame with its pre-computed Metric3D \citep{yin2023metric3d} depth; the frozen DINOv2 backbone is timed through its \texttt{\detokenize{forward_features}} call. GaussTR is the one model in this study whose Gaussian-side cost dominates: every query becomes a Gaussian per view and the voxelizer splats all of them into the $200\times200\times16$ grid, so removing 47\% of the queries cuts the voxelizer by 45\% and the whole model by 17.8\%. This saving is what any 47\% reduction of the query bank buys; because the ranking has no signal, it is the effect of choosing a smaller bank, not of Decision-FLIP, and it comes with a 0.81-point (6.6\% relative) mIoU loss with no recovery. We therefore report it as a boundary case and not as a compression result.

\subsection{Cross-architecture qualitative results}\label{sec:qualitative-results}

\begin{figure}[t]
\centering
\includegraphics[width=\linewidth]{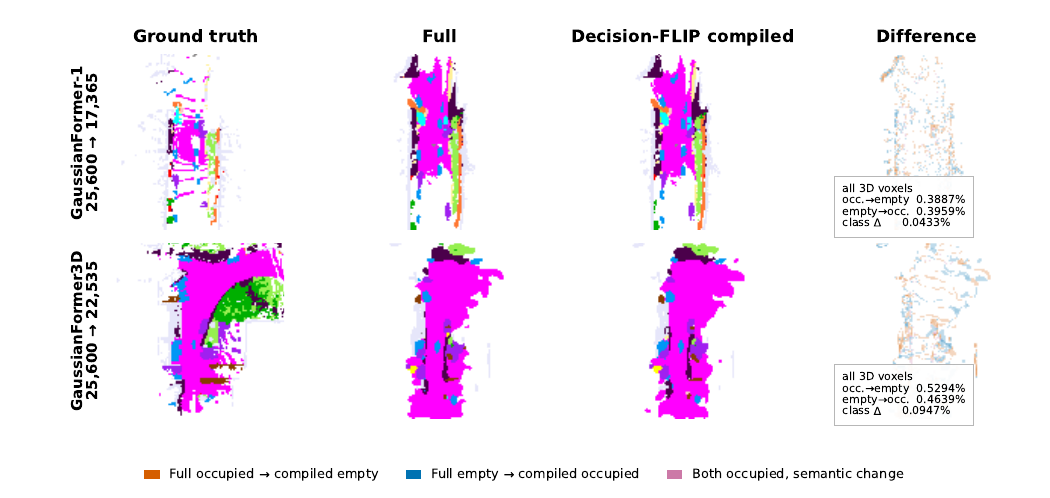}
\caption{Cross-architecture SurroundOcc qualitative results. Rows show the primary endpoints after matched recovery for GaussianFormer-1 (full $K=25{,}600$, compiled $K=17{,}365$) and GaussianFormer3D (full $K=25{,}600$, compiled $K=22{,}535$). Columns are ground truth (GT), Full, Decision-FLIP Compiled, and Difference. Difference percentages use all 640,000 voxels in each 200$\times$200$\times$16 grid as the denominator.}
\label{fig:qualitative-surroundocc}
\end{figure}

\begin{figure}[t]
\centering
\includegraphics[width=\linewidth]{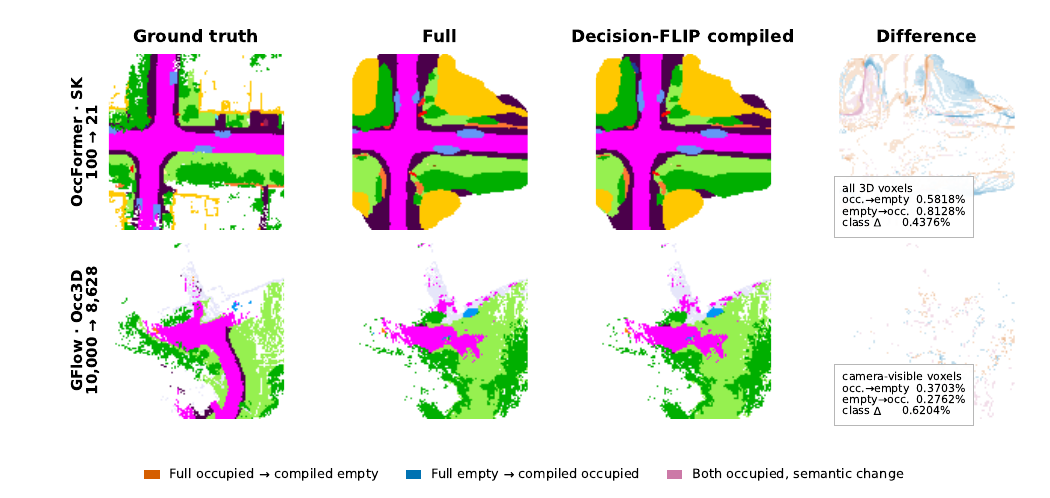}
\caption{Cross-architecture qualitative results for the exact primary matched-recovery endpoints: OccFormer on SemanticKITTI (full $K=100$, compiled $K=21$) and GaussianFlowOcc on Occ3D-nuScenes~\citep{tian2023occ3d} (full $K=10{,}000$, guarded-Otsu compiled $K=8{,}628$, $\tau=0.2$). Both samples were fixed before inspecting model output. Columns are ground truth (GT), Full, Decision-FLIP Compiled, and Difference. OccFormer percentages use all 2,097,152 voxels in its 256$\times$256$\times$32 grid; GaussianFlowOcc percentages use the 130,720 camera-visible voxels in its 200$\times$200$\times$16 grid.}
\label{fig:qualitative-semkitti-occ3d}
\end{figure}

The samples in Figures~\ref{fig:qualitative-surroundocc} and~\ref{fig:qualitative-semkitti-occ3d} were selected by a fixed deterministic rule over each complete validation set before inspecting any model output.

\subsection{Full per-class results}\label{sec:perclass-results}

This section reports every semantic class for the compiled checkpoints, including changes on rare categories. All entries are IoU percentages. The benchmark-specific tables pair the matched full model after one recovery epoch with the adaptive compiled model after the same epoch, grouped by benchmark because the class ontologies differ. Boldface identifies the compiled column for visual grouping; it does not claim that the compiled value is better. All evaluations use the complete evaluation set defined in Appendix~\ref{sec:protocols}. Lifecycle Mean-FLIP reduces exactly to Mean-FLIP for the always-alive GF1, GF3D, OccFormer, and GaussianFlowOcc banks; GaussianWorld uses its streaming lifecycle integral. Endpoint choices that differ between models are given in the table notes.

\begin{table}[t]
\caption{Per-class preservation across architectures}\label{tab:main-perclass-summary}
\centering
\small
\setlength{\tabcolsep}{3pt}
\begin{tabular}{lrr}
\toprule
\textbf{Model} & \textbf{Largest class drop} & \textbf{Largest upward change} \\
\midrule
GaussianWorld & vegetation -0.26 & trailer +0.15 \\
GaussianFormer-1 & vegetation -0.40 & none (best: construction vehicle -0.02) \\
GaussianFormer3D & manmade -0.18 & other flat +0.16 \\
OccFormer (paired evaluation) & other-vehicle -0.59 & terrain +0.72 \\
GaussianFlowOcc & trailer -0.33 & barrier +0.32 \\
\bottomrule
\end{tabular}%
\par\vspace{3pt}\begin{minipage}{\linewidth}\footnotesize Statistics use every official semantic class after matched recovery; changes are compiled minus matched full in class-IoU points. All five primary matched pairs have complete per-class outputs.\end{minipage}
\end{table}

\begin{table}[t]
\caption{SurroundOcc per-class preservation after matched recovery}\label{tab:perclass-surroundocc}
\centering
\footnotesize
\setlength{\tabcolsep}{3pt}
\fitwidth{%
\begin{tabular}{lrrrrrr}
\toprule
\textbf{Metric / class} & \textbf{GW full} & \textbf{GW + ours} & \textbf{GF1 full} & \textbf{GF1 + ours} & \textbf{GF3D full} & \textbf{GF3D + ours} \\
\midrule
Geometry IoU & 33.097 & 32.978 & 30.049 & 29.808 & 35.496 & 35.380 \\
barrier & 21.77 & 21.57 & 19.64 & 19.44 & 18.91 & 18.96 \\
bicycle & 13.56 & 13.45 & 12.44 & 12.29 & 11.99 & 12.05 \\
bus & 26.86 & 26.72 & 25.74 & 25.52 & 28.03 & 28.10 \\
car & 31.29 & 31.07 & 29.41 & 29.37 & 31.37 & 31.40 \\
construction vehicle & 13.72 & 13.47 & 10.40 & 10.38 & 15.01 & 14.95 \\
motorcycle & 16.79 & 16.91 & 15.50 & 15.12 & 16.93 & 17.01 \\
pedestrian & 13.24 & 13.23 & 12.80 & 12.69 & 16.26 & 16.10 \\
traffic cone & 11.44 & 11.51 & 8.93 & 8.84 & 5.81 & 5.78 \\
trailer & 15.03 & 15.18 & 12.82 & 12.43 & 15.96 & 15.82 \\
truck & 23.65 & 23.50 & 20.73 & 20.55 & 24.63 & 24.56 \\
driveable surface & 42.23 & 42.20 & 39.14 & 39.10 & 38.06 & 38.05 \\
other flat & 24.69 & 24.65 & 21.82 & 21.77 & 18.12 & 18.28 \\
sidewalk & 28.69 & 28.57 & 24.49 & 24.37 & 23.24 & 23.25 \\
terrain & 26.29 & 26.32 & 23.05 & 22.84 & 22.57 & 22.51 \\
manmade & 15.64 & 15.51 & 11.57 & 11.41 & 19.69 & 19.51 \\
vegetation & 25.08 & 24.82 & 20.22 & 19.82 & 31.83 & 31.72 \\
\textbf{Semantic mIoU} & \textbf{21.874} & \textbf{21.792} & \textbf{19.293} & \textbf{19.121} & \textbf{21.152} & \textbf{21.128} \\
\bottomrule
\end{tabular}%
}
\par\vspace{3pt}\begin{minipage}{\linewidth}\footnotesize \textbf{Note.} GF1 columns are the seed-0 matched-recovery pair, with aggregates 19.293 / 30.049 (full) versus 19.121 / 29.808 (compiled); they are not the three-seed aggregate in Appendix~\ref{sec:matched-recovery-gf1}. GF3D uses one fixed matched-recovery run.\end{minipage}
\end{table}

\begin{table}[t]
\caption{Occ3D-nuScenes~\citep{tian2023occ3d} per-class preservation for GaussianFlowOcc after matched recovery}\label{tab:perclass-occ3d}
\centering
\footnotesize
\setlength{\tabcolsep}{3pt}
\fitwidth{%
\begin{tabular}{lrrr}
\toprule
\textbf{Metric / class} & \textbf{Full $K=10{,}000$} & \textbf{Guarded-Otsu $K=8{,}628$} & \textbf{Change} \\
\midrule
Occupied IoU & 54.75 & \textbf{54.78} & +0.03 \\
barrier & 24.75 & \textbf{25.07} & +0.32 \\
bicycle & 9.16 & \textbf{9.24} & +0.08 \\
bus & 24.85 & \textbf{24.85} & +0.00 \\
car & 31.32 & \textbf{31.55} & +0.23 \\
construction vehicle & 13.06 & \textbf{13.17} & +0.11 \\
motorcycle & 10.33 & \textbf{10.19} & -0.14 \\
pedestrian & 12.91 & \textbf{12.86} & -0.05 \\
traffic cone & 8.64 & \textbf{8.45} & -0.19 \\
trailer & 17.99 & \textbf{17.66} & -0.33 \\
truck & 21.22 & \textbf{21.21} & -0.01 \\
driveable surface & 73.42 & \textbf{73.47} & +0.05 \\
sidewalk & 42.39 & \textbf{42.48} & +0.09 \\
terrain & 45.41 & \textbf{45.39} & -0.02 \\
manmade & 21.49 & \textbf{21.50} & +0.01 \\
vegetation & 13.55 & \textbf{13.50} & -0.05 \\
\textbf{Semantic mIoU} & \textbf{24.70} & \textbf{24.71} & \textbf{+0.01} \\
\bottomrule
\end{tabular}%
}
\par\vspace{3pt}\begin{minipage}{\linewidth}\footnotesize \textbf{Note.} Both columns are the exact matched-recovery endpoints evaluated at $\tau=0.2$ on all 6,019 validation frames. The repository evaluator averages the 15 classes shown (it excludes \texttt{\detokenize{others}} and \texttt{\detokenize{other flat}}). Checkpoint and evaluation-log identifiers will be released with the code.\end{minipage}
\end{table}

Across the five primary settings with complete per-class outputs, no class that is nonzero in its matched full model newly collapses to zero after adaptive compilation and matched recovery. The largest drop is 0.59 points (OccFormer other-vehicle); GaussianFlowOcc's largest drop is 0.33 points (trailer), and its largest gain is 0.32 points (barrier).

Table~\ref{tab:perclass-provenance} adds the recovery stage for GaussianWorld and GaussianFormer-1: each class is shown before and after the recovery epoch, so the reader can see how much of the matched gap recovery closes and verify that no class that is nonzero in the matched full model newly collapses to zero.

\begin{table}[t]
\caption{Recovery-stage per-class results for GaussianWorld and GaussianFormer-1}\label{tab:perclass-provenance}
\centering
\footnotesize
\setlength{\tabcolsep}{3pt}
\fitwidth{%
\begin{tabular}{lrrrrrr}
\toprule
\textbf{Metric / class} & \textbf{GW full} & \textbf{GW static reference (0 ep)} & \textbf{GW static reference (1 ep)} & \textbf{GF1 full} & \textbf{GF1 adaptive (0 ep)} & \textbf{GF1 adaptive (1 ep)} \\
\midrule
Geometry IoU & 33.097$\pm$0.053 & 32.566 & \textbf{32.827} & 30.049 & 28.344 & \textbf{29.808} \\
barrier & 21.77 & 21.58 & 21.68 & 19.64 & 17.92 & \textbf{19.44} \\
bicycle & 13.56 & 12.92 & 13.57 & 12.44 & 10.63 & \textbf{12.29} \\
bus & 26.86 & 26.83 & 26.81 & 25.74 & 21.57 & \textbf{25.52} \\
car & 31.29 & 30.94 & 30.80 & 29.41 & 26.23 & \textbf{29.37} \\
construction vehicle & 13.72 & 13.81 & 13.60 & 10.40 & 9.57 & \textbf{10.38} \\
motorcycle & 16.79 & 16.62 & 16.61 & 15.50 & 13.41 & \textbf{15.12} \\
pedestrian & 13.24 & 13.11 & 13.11 & 12.80 & 11.35 & \textbf{12.69} \\
traffic cone & 11.44 & 11.28 & 11.15 & 8.93 & 7.61 & \textbf{8.84} \\
trailer & 15.03 & 14.68 & 14.72 & 12.82 & 11.85 & \textbf{12.43} \\
truck & 23.65 & 23.47 & 23.31 & 20.73 & 18.59 & \textbf{20.55} \\
driveable surface & 42.23 & 41.93 & 41.98 & 39.14 & 36.39 & \textbf{39.10} \\
other flat & 24.69 & 24.49 & 24.37 & 21.82 & 19.40 & \textbf{21.77} \\
sidewalk & 28.69 & 28.34 & 28.50 & 24.49 & 22.67 & \textbf{24.37} \\
terrain & 26.29 & 25.83 & 26.13 & 23.05 & 21.88 & \textbf{22.84} \\
manmade & 15.64 & 15.15 & 15.41 & 11.57 & 11.38 & \textbf{11.41} \\
vegetation & 25.08 & 24.65 & 24.63 & 20.22 & 18.17 & \textbf{19.82} \\
Semantic mIoU & 21.874$\pm$0.025 & 21.490 & \textbf{21.649} & 19.293 & 17.414 & \textbf{19.121} \\
\bottomrule
\end{tabular}%
}
\par\vspace{3pt}\begin{minipage}{\linewidth}\footnotesize \textbf{Note.} The comparison is matched full one-epoch recovery versus adaptive compiled one-epoch recovery; the "(0 ep)" columns give the same compiled checkpoints before recovery. GW columns retain the static $K=16{,}461$ reference; the lifecycle result is reported in Appendix~\ref{sec:gw-results} and Table~\ref{tab:compiled-vs-full}. GF1 uses the seed-0 matched-recovery pair, and all GaussianFormer values cover 753 validation batches.\end{minipage}
\end{table}

Table~\ref{tab:occformer-detail} is the primary recovered per-class comparison for OccFormer; both arms use the same one-epoch recipe and cover all 815 validation frames.

\begin{table}[t]
\caption{OccFormer per-class sensitivity to budget and recovery recipe}\label{tab:occformer-detail}
\centering
\footnotesize
\setlength{\tabcolsep}{3pt}
\fitwidth{%
\begin{tabular}{lrr}
\toprule
\textbf{Class} & \textbf{Matched full $K=100$} & \textbf{Otsu $K=21$} \\
\midrule
car & 24.49 & 24.89 \\
bicycle & 1.39 & 1.32 \\
motorcycle & 1.44 & 1.10 \\
truck & 17.75 & 17.51 \\
other-vehicle & 12.15 & 11.56 \\
person & 3.27 & 3.17 \\
bicyclist & 4.34 & 3.94 \\
motorcyclist & 0.00 & 0.00 \\
road & 58.99 & 58.89 \\
parking & 18.96 & 18.93 \\
sidewalk & 25.02 & 25.66 \\
other-ground & 0.30 & 0.35 \\
building & 14.49 & 14.65 \\
fence & 5.15 & 5.33 \\
vegetation & 19.50 & 19.52 \\
trunk & 3.57 & 3.63 \\
terrain & 29.49 & 30.21 \\
pole & 3.96 & 3.97 \\
traffic-sign & 2.63 & 2.71 \\
\textbf{SSC mIoU} & \textbf{12.99} & \textbf{13.02} \\
\textbf{SC IoU} & \textbf{36.22} & \textbf{36.30} \\
\bottomrule
\end{tabular}%
}
\par\vspace{3pt}\begin{minipage}{\linewidth}\footnotesize \textbf{Primary paired evaluation.} Compiled minus full is +0.03 SSC mIoU / +0.08 SC IoU. Other-vehicle has the largest class drop (-0.59); terrain has the largest increase (+0.72); no nonzero full class newly collapses.\end{minipage}
\end{table}

Table~\ref{tab:perclass-released} lists evaluator anchors. Its columns anchor the provenance chain (released $\rightarrow$ source $\rightarrow$ compiled) and establish evaluator comparability only; they are not matched controls, and no pruning conclusion rests on them.

\begin{table}[t]
\caption{Released-checkpoint per-class evaluator anchors}\label{tab:perclass-released}
\centering
\footnotesize
\setlength{\tabcolsep}{3pt}
\fitwidth{%
\begin{tabular}{lrr}
\toprule
\textbf{Metric / class} & \textbf{GW released \texttt{\detokenize{ckpt_stream}}} & \textbf{GF1 released solid} \\
\midrule
Geometry IoU & 32.804 & 28.779 \\
barrier & 21.47 & 17.86 \\
bicycle & 13.53 & 10.60 \\
bus & 27.19 & 22.43 \\
car & 31.18 & 26.08 \\
construction vehicle & 13.77 & 9.49 \\
motorcycle & 16.20 & 13.35 \\
pedestrian & 13.31 & 11.31 \\
traffic cone & 11.85 & 7.55 \\
trailer & 14.85 & 12.08 \\
truck & 23.45 & 18.73 \\
driveable surface & 42.00 & 37.02 \\
other flat & 24.49 & 19.19 \\
sidewalk & 28.44 & 22.84 \\
terrain & 26.32 & 22.02 \\
manmade & 15.60 & 11.90 \\
vegetation & 24.49 & 18.69 \\
Semantic mIoU & 21.759 & 17.571 \\
\bottomrule
\end{tabular}%
}
\par\vspace{3pt}\begin{minipage}{\linewidth}\footnotesize \textbf{Note.} The same architecture-specific evaluators are used throughout Appendix A.\end{minipage}
\end{table}

\subsection{Full compilation and matched-recovery table}

Table~\ref{tab:compiled-vs-full} gives one primary row per eligible setting: the full and adaptive compiled checkpoints before recovery, then the same pair after matched one-epoch recovery. The metric pair and benchmark are explicit in every row, and every number is a percentage. Differences are compiled minus full. Model-specific matched-recovery evidence remains in the architecture tables, with the GaussianFormer-1 seed control in Appendix~\ref{sec:matched-recovery-gf1}.

\begin{table}[t]
\caption{Model-dataset compilation and matched-recovery results}\label{tab:compiled-vs-full}
\centering
\footnotesize
\setlength{\tabcolsep}{3pt}
\fitwidth{%
\begin{tabular}{lrrrrrrr}
\toprule
\textbf{Model (benchmark; metric pair)} & \textbf{Eligible $K\rightarrow$ kept (removed)} & \textbf{Full, 0 ep} & \textbf{Ours, 0 ep} & \textbf{$\Delta$ 0 ep} & \textbf{Full, matched 1 ep} & \textbf{Ours, matched 1 ep} & \textbf{$\Delta$ 1 ep} \\
\midrule
GaussianWorld (SurroundOcc; mIoU / geometry IoU) & 25,600 $\rightarrow$ 22,427 (12.4\%) & 21.902 / 33.140 & 21.723 / 32.850 & -0.179 / -0.290 & 21.874$\pm$0.025 / 33.097$\pm$0.053 & 21.792 / 32.978 & -0.082 / -0.119 \\
GaussianFormer-1 (SurroundOcc; mIoU / geometry IoU) & 25,600 $\rightarrow$ 17,365 (32.2\%) & 17.571 / 28.779 & 17.414 / 28.344 & -0.157 / -0.435 & 19.291$\pm$0.002 / 30.050$\pm$0.002 & 19.120$\pm$0.003 / 29.804$\pm$0.001 & -0.171$\pm$0.002 / -0.246$\pm$0.001 \\
GaussianFormer3D (SurroundOcc; mIoU / geometry IoU) & 25,600 $\rightarrow$ 22,535 (12.0\%) & 18.400 / 30.056 & 18.356 / 29.963 & -0.044 / -0.093 & 21.152 / 35.496 & 21.128 / 35.380 & -0.024 / -0.116 \\
OccFormer (SemanticKITTI; SSC mIoU / SC IoU) & 100 $\rightarrow$ 21 (79.0\%) & 13.50 / 36.42 & 13.34 / 36.42 & -0.16 / 0.00 & 12.99 / 36.22 & 13.02 / 36.30 & +0.03 / +0.08 \\
GaussianFlowOcc (Occ3D; voxel mIoU / occupied IoU) & 10,000 $\rightarrow$ 8,628 (13.72\%) & 24.65 / 54.86 & 24.75 / 55.43 & +0.10 / +0.57 & 24.70 / 54.75 & 24.71 / 54.78 & +0.01 / +0.03 \\
\bottomrule
\end{tabular}%
}
\end{table}

Two regularities hold across the eligible model-dataset settings. First, automatic compilation preserves accuracy where the bank has stable row identity: before recovery, semantic mIoU falls by at most 0.179 point and the second reported metric by at most 0.435 point; GaussianFlowOcc instead improves on both metrics. Second, after matched recovery every semantic gap is at most 0.171 point and every geometry/occupancy gap is at most 0.246 point. GaussianFlowOcc remains 0.03 occupied-IoU point above its full control. Where recovery changes the absolute level substantially (GaussianFormer-1 +1.7 mIoU and GaussianFormer3D +2.7), the shift is pipeline adaptation shared by both arms, which is why only matched pairs are compared.

\subsection{Full ablations and diagnostics}\label{sec:diagnostics}

\subsubsection{Same-\texorpdfstring{$K$}{K} ranking quality before recovery}

The same-$K$ comparison fixes each architecture at the survivor count selected by its adaptive Exact Mean-FLIP/Otsu arm and changes only the ranking. Every number is from the same source checkpoint and complete official validation; no arm receives recovery. This isolates score quality from both budget selection and fine-tuning. Unavailable entries denote controls that are undefined or have not completed under the same strict protocol.

\begin{table}[t]
\caption{Same-$K$ ranking controls before recovery}\label{tab:samek-full}
\centering
\footnotesize
\setlength{\tabcolsep}{3pt}
\fitwidth{%
\begin{tabular}{lrrrrrrrr}
\toprule
\textbf{Model (metric pair)} & \textbf{$K$} & \textbf{Full} & \textbf{Exact Mean-FLIP} & \textbf{Taylor} & \textbf{Random} & \textbf{Fire/activity} & \textbf{Opacity} & \textbf{Magnitude/query norm} \\
\midrule
GaussianWorld (mIoU / geometry IoU) & 22,427 & 21.902 / 33.140 & \textbf{21.723 / 32.850} & 21.708 / 32.762 & 21.512 / 32.342 & 21.650 / 32.665 & 21.658 / 32.699 & 21.576 / 32.388 \\
GaussianFormer-1 (mIoU / geometry IoU) & 17,365 & 17.571 / 28.779 & \textbf{17.414 / 28.344} & 16.349 / 25.961 & 9.261 / 11.809 & 8.919 / 11.582 & 17.388 / 27.566 & 17.097 / 27.319 \\
GaussianFormer3D (mIoU / geometry IoU) & 22,535 & 18.400 / 30.056 & \textbf{18.356 / 29.963} & 18.325 / 29.873 & 17.67 / 28.44 & 18.30 / 29.86 & 18.295 / 29.915 & 18.329 / 29.843 \\
OccFormer (SSC mIoU / SC IoU) & 21 & 13.50 / 36.42 & \textbf{13.34 / 36.42} & 5.52 / 7.88 & 0.82 / 7.89 & 13.17 / 36.41 & --- & 1.53 / 7.88 \\
GaussianFlowOcc, primary sequential guarded Otsu (voxel mIoU / occupied IoU) & 8,628 & 24.65 / 54.86 & \textbf{24.75 / 55.43} & 24.75 / 55.41 & unavailable & unavailable & unavailable & unavailable \\
\bottomrule
\end{tabular}%
}
\end{table}

Exact Mean-FLIP matches or exceeds supervised Taylor on the primary semantic metric in all five settings. The largest margins occur on GaussianFormer-1 and OccFormer; GaussianFlowOcc ties in mIoU, with Mean-FLIP 0.02 point higher in occupied IoU. Random selection collapses GaussianFormer-1 and OccFormer. Thus the completed same-$K$ results rule out the explanation that any cut of the selected size works. GaussianFlowOcc's other primary $K=8{,}628$ proxy controls are unavailable.

Matched recovery controls answer a different question: whether fine-tuning can erase ranking differences. At the GaussianFormer-1 fixed comparison budget $K=17{,}296$ (Appendix~\ref{sec:gf1-results}), recovery repairs part, but not all, of the random cut: the random arm ends 0.6 mIoU below the exact and opacity arms after the same epoch. We do not extrapolate this result to architectures whose matched recovered proxy arms have not completed.

\subsubsection{Decision criterion and thresholding ablation}

The exact leave-one-out score can count changed decisions (hard FLIP, the criterion used throughout) or accumulate the total-variation distance between the full and leave-one-out class distributions (TV-FLIP, $s_i^{TV}=\tfrac12\sum_x\lVert p_f(x)-p_{f,-i}(x)\rVert_1$); and the Otsu split can be applied once or re-applied to the low mode after each removal round (iterative Otsu). We ran all four combinations on the released GaussianFormer-1 checkpoint with 100 training calibration frames and 50 disjoint held-out training frames, and report the exact set-removal drift of the compiled readout on held-out frames (fraction of voxels the full model predicts occupied whose argmax changes after removing the whole set), with matched-$K$ masks so that every method is read at every budget. This independent 100-frame criterion subset is distinct from the nested 100-frame prefix used in the calibration-size sweep; consequently, their selected $K$ values are not expected to coincide.

\begin{table}[t]
\caption{Hard and total-variation FLIP with single-shot and iterative Otsu}\label{tab:criterion}
\centering
\footnotesize
\setlength{\tabcolsep}{3pt}
\fitwidth{%
\begin{tabular}{lrrr}
\toprule
\textbf{Held-out occupied-voxel argmax drift} & \textbf{$K=19{,}830$ (TV Otsu)} & \textbf{$K=18{,}769$ (hard Otsu)} & \textbf{$K=14{,}328$ (hard, iteration 2)} \\
\midrule
Hard FLIP, single-shot & \textbf{0.040} & \textbf{0.055} & \textbf{0.171} \\
Hard FLIP, iterative Otsu & 0.144 & 0.148 & 0.164 \\
TV-FLIP, single-shot & 0.239 & 0.296 & 0.516 \\
TV-FLIP, iterative Otsu & 0.515 & 0.555 & 0.798 \\
Random ranking & 0.596 & 0.669 & 0.879 \\
Fire-rate ranking & 0.596 & 0.669 & 0.879 \\
\bottomrule
\end{tabular}%
}
\end{table}

Random and fire-rate coincide in this evaluation because all slots exceed the fire threshold on the independent 100-frame subset, leaving fire-rate non-discriminative; the saved masks and resulting drifts are therefore identical.

Both scores are reliable rankings (split-half Spearman 0.925 hard, 0.929 TV) but they rank different slots (Spearman 0.39 between them): TV keeps diffuse slots that move probability mass without changing decisions and drops small decisive ones, so at every budget it is worse on the quantity compilation must preserve. Iterative re-thresholding does not improve the ranking (at $K=14{,}328$ it reads 0.164 against 0.171 for single-shot on this one held-out split; no variance estimate is available); it only cuts deeper, and the separation gate does not stop it: the TV variant ran eight rounds down to $K=76$ with $\eta$ between 0.55 and 0.70 at every round. The separation gate is therefore a plausibility check on one split, not a stopping rule, and the budget of Section~\ref{sec:otsu} is the single-shot valley. The remaining 5.5\% drift at the hard Otsu budget is a joint-removal cost that no re-ranking removes; it is the cost the matched recovery epoch addresses.

\subsubsection{Label supervision and self-distillation during recovery}

The compiled model is selected to preserve the full model's decisions, whereas the recovery epoch fits ground-truth labels. To test whether the residual gap of the static GaussianWorld reference is an artifact of that mismatch, we repeated recovery with a frozen full-bank teacher (the calibrated source checkpoint, run in lock-step through the same 38-frame streams) and a per-voxel KL term between teacher and student voxel logits (weight 100, temperature 1), either added to the task loss or replacing it, with the unpruned bank receiving the identical recipe as control. Everything else (one epoch, seed, learning rate, evaluator) matches the static reference.

\begin{table}[t]
\caption{Label supervision and self-distillation during recovery}\label{tab:recovery-objective}
\centering
\footnotesize
\setlength{\tabcolsep}{3pt}
\fitwidth{%
\begin{tabular}{lrrr}
\toprule
\textbf{GaussianWorld, $K=16{,}461$, one epoch, 5,700 frames} & \textbf{Full} & \textbf{Compiled} & \textbf{$\Delta$} \\
\midrule
Labels (static-reference recipe) & 21.874 / 33.097 & 21.649 / 32.827 & -0.225 / -0.270 \\
Labels + self-distillation & 21.843 / 32.870 & 21.586 / 32.489 & -0.257 / -0.381 \\
Self-distillation only (label-free) & --- & 21.581 / 32.420 & --- \\
\bottomrule
\end{tabular}%
}
\end{table}

Adding the KL term does not close the gap: both arms end slightly below the label-trained ones (the geometry IoU by 0.2 to 0.4), the compiled-versus-full difference is unchanged (-0.26 / -0.38 against -0.23 / -0.27), and the label-free and mixed variants end within 0.005 mIoU on this run. The residual cost of the 36\% cut is therefore not a property of the recovery objective.

\subsubsection{Recovery interventions at fixed GaussianWorld budgets}

Earlier exploratory runs on the same evaluator varied the recovery itself at fixed budgets (Mean-FLIP ranking, one epoch unless stated; 5,700 frames); Table~\ref{tab:recovery-levers} lists them.

\begin{table}[t]
\caption{Recovery interventions at fixed GaussianWorld budgets}\label{tab:recovery-levers}
\centering
\footnotesize
\setlength{\tabcolsep}{3pt}
\fitwidth{%
\begin{tabular}{lrr}
\toprule
\textbf{Lever (GaussianWorld)} & \textbf{$K$} & \textbf{mIoU / IoU} \\
\midrule
Standard recovery, three seeds & 13,364 & 21.28 / 32.29, 21.38 / 32.41, 21.36 / 32.38 \\
Anchor-position learning rate $\times$4 & 13,364 & 21.37 / 32.34 \\
Standard recovery & 20,271 & 21.72 / 32.99 (compiled before recovery: 21.73 / 32.96) \\
Relocating instead of deleting: the 5,329 lowest-scored anchors are moved, none removed & 25,600 & 21.31 / 32.42 before recovery, 21.70 / 32.94 after \\
Progressive pruning in three rounds, last round & 13,364 & 21.15 / 32.07 \\
Three recovery epochs (uncalibrated source) & 13,364 & 20.83 $\rightarrow$ 21.24 $\rightarrow$ 21.26 after epochs 1, 2, 3 \\
\bottomrule
\end{tabular}%
}
\end{table}

None of these interventions improves the endpoint. A larger anchor learning rate and longer training land where standard recovery lands, progressive pruning lands below it, and at the 21\% budget recovery changes nothing at all. Relocating the low-scored anchors instead of deleting them keeps the full bank's cost yet ends 0.17 mIoU below the recovered full control, so moving idle slots is not a free alternative to removing them. Together with the criterion ablation above, this places the residual cost with the capacity of the smaller bank rather than with the optimization of the recovery.

\subsubsection{Label-free empty-vote recalibration}

The GaussianWorld head adds one fixed empty Gaussian whose vote into the empty class scales with a learned scalar. Compilation removes scene Gaussians but not this vote, so voxels that lost part of their mass tip towards empty; this is why the zero-recovery loss is concentrated in geometry (the $\Delta$ 0 ep column of Table~\ref{tab:compiled-vs-full}). Because the empty vote enters the argmax linearly, rescaling the scalar by $\alpha$ is exact and can be replayed from one head pass. We chose $\alpha$ label-free on 100 training frames so that the compiled model's occupied-voxel rate matches the full model's, and read the result on 50 held-out training frames under the local single-frame protocol (absolute values are not comparable with the streaming evaluation). Table~\ref{tab:recalibration} reports the result.

\begin{table}[t]
\caption{Label-free empty-vote recalibration}\label{tab:recalibration}
\centering
\footnotesize
\setlength{\tabcolsep}{3pt}
\fitwidth{%
\begin{tabular}{lr}
\toprule
\textbf{GaussianWorld, $K=16{,}461$, 0 recovery, 50 held-out train frames} & \textbf{mIoU / IoU} \\
\midrule
Full bank & 0.2803 / 0.4476 \\
Compiled, $\alpha=1$ & 0.2761 / 0.4244 \\
Compiled, $\alpha^\ast=0.976$ (occupancy-matched) & 0.2746 / 0.4316 \\
Compiled, $\alpha=0.95$ & 0.2726 / 0.4359 \\
\bottomrule
\end{tabular}%
}
\end{table}

The single global rescale returns one third to one half of the geometry loss and costs a little mIoU, because the missing vote mass is spatially uneven: voxels whose Gaussians were removed lose more than a global factor restores. It is a partial, recovery-free fix, not a substitute for recovery. Per-Gaussian opacity compensation (a least-squares reweighting of the survivors' opacities, also label-free) is the natural extension and is left for future work.

\subsubsection{Automatic-selector comparison}\label{sec:selector-comparison}

Each selector operates on the same fixed Mean-FLIP evidence within an architecture and chooses its own $K$ without validation metrics. The cross-architecture comparison uses zero-recovery semantic mIoU relative to each model's evaluated full source checkpoint, avoiding differences among recovery recipes.

\begin{table}[t]
\caption{Cross-architecture automatic-selector operating points before recovery}\label{tab:selector-cross-model}
\centering
\footnotesize
\setlength{\tabcolsep}{3pt}
\fitwidth{%
\begin{tabular}{lrrrr}
\toprule
\textbf{Model} & \textbf{Guarded Otsu (removed / $\Delta$mIoU)} & \textbf{Two-component GMM (removed / $\Delta$mIoU)} & \textbf{Kneedle (removed / $\Delta$mIoU)} & \textbf{Largest-gap (removed / $\Delta$mIoU)} \\
\midrule
GaussianWorld & 12.4\% / -0.179 & 22.0\% / -0.544 & 12.4\% / -0.179 & 12.4\% / -0.179 \\
GaussianFormer-1 & 32.2\% / -0.157 & 45.2\% / -0.552 & 99.6\% / -17.531 & 0.035\% / -0.001 \\
GaussianFormer3D & 12.0\% / -0.044 & 9.8\% / -0.030 & 99.4\% / -17.940 & 2.4\% / -0.006 \\
GaussianFlowOcc & 13.7\% / +0.10 & 41.1\% / -0.24 & 12.8\% / +0.09 & 12.8\% / +0.09 \\
OccFormer (SemanticKITTI) & 79\% / -0.16 & 63\% / 0.00 & 76\% / -0.02 & 3\% / +0.06 \\
\bottomrule
\end{tabular}%
}
\end{table}

GaussianWorld's Otsu, Kneedle, and largest-gap rules produce the same physical mask; Kneedle and largest-gap likewise coincide on GaussianFlowOcc. The remaining differences illustrate the operating-point trade-off: conservative rules can preserve accuracy by removing almost nothing, whereas aggressive rules can collapse the model or incur a larger loss.

The detailed GaussianFormer-1 comparison fixes its saved 500-frame Mean-FLIP vector and per-frame matrix before any validation result is read. Each non-abstaining mask is physically compiled, loaded under strict shape and key checking, evaluated on all 753 official validation batches, recovered for one matched epoch, and evaluated again. The common full matched-recovery control is 19.293 mIoU / 30.049 geometry IoU.

\begin{table}[t]
\caption{GaussianFormer-1 recovered-selector detail}\label{tab:selector-ablation}
\centering
\footnotesize
\setlength{\tabcolsep}{3pt}
\fitwidth{%
\begin{tabular}{lrrrrrrr}
\toprule
\textbf{Selector} & \textbf{Status} & \textbf{$K$ (removed)} & \textbf{Split-half $K_A/K_B$} & \textbf{Mask Jaccard} & \textbf{Before recovery mIoU / geometry IoU} & \textbf{After matched recovery mIoU / geometry IoU} & \textbf{Recovered $\Delta$ vs. full} \\
\midrule
Guarded Otsu & selected & 17,365 (32.168\%) & 17,442 / 18,019 & 0.928 & 17.413 / 28.344 & 19.120 / 29.804 & -0.173 / -0.245 \\
Two-component GMM & selected & 14,035 (45.176\%) & 14,261 / 15,750 & 0.890 & 17.019 / 27.313 & 18.928 / 29.543 & -0.365 / -0.507 \\
Kneedle & selected, degenerate operating point & 92 (99.641\%) & 103 / 114 & 0.428 & 0.040 / 0.034 & 2.076 / 2.352 & -17.217 / -27.698 \\
Largest-gap & selected & 25,591 (0.035\%) & 25,577 / 25,563 & 0.998 & 17.570 / 28.779 & 19.290 / 30.051 & -0.003 / +0.002 \\
\bottomrule
\end{tabular}%
}
\end{table}

Largest-gap removes only the nine rows with exactly zero Mean-FLIP ($K=25{,}591$). Identifiers for the mask, score vector, frame matrix, pinned full-control metrics, and aggregate summary are recorded in a provenance manifest that will be released with the code. An independent run produced byte-identical masks and complete 753-batch evaluations. No validation metric participates in selection. The comparison is not matched in $K$, and it covers one checkpoint, calibration set, and recovery seed, so it establishes an operating-point trade-off rather than universal selector optimality.

\begin{figure}[t]
\centering
\includegraphics[width=\linewidth]{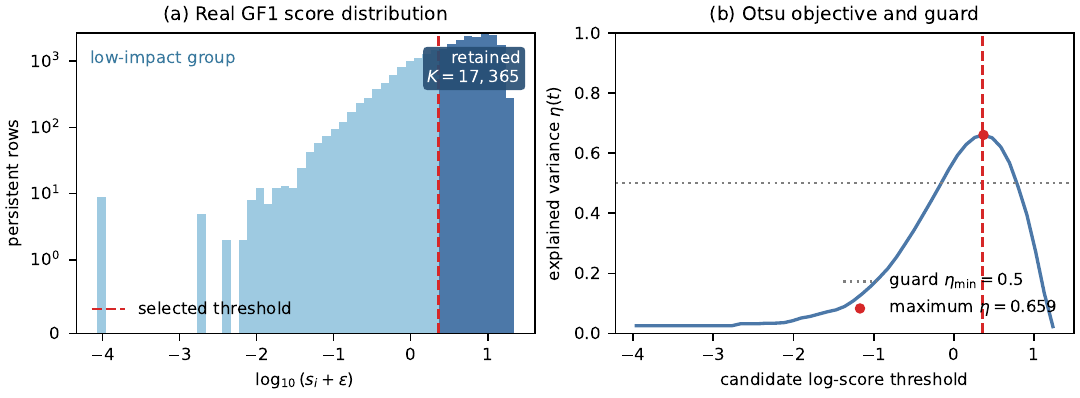}
\caption{Detailed guarded-Otsu diagnostic on GaussianFormer-1. Left: the saved log Mean-FLIP distribution and selected partition. Right: normalized between-group variance over candidate thresholds; guarded Otsu compiles only when the maximizing separation exceeds the fixed gate. This diagnostic is retained as distributional evidence and does not imply universal bimodality or accuracy optimality.}
\label{fig:otsu-diagnostic}
\end{figure}

\subsubsection{Separation and calibration diagnostics}

The identity eligibility check and score-stability check serve different purposes. Identity is an architectural or lineage property (Section~\ref{sec:banks}); split-half statistics ask whether a finite calibration sample estimates a stable ranking or threshold. Table~\ref{tab:separation} lists the diagnostics; Figure~\ref{fig:otsu-diagnostic} plots the GaussianFormer-1 log-score distribution and Otsu separation curve.

\begin{table}[t]
\caption{Separation and calibration diagnostics}\label{tab:separation}
\centering
\footnotesize
\setlength{\tabcolsep}{3pt}
\fitwidth{%
\begin{tabular}{lrrrrrr}
\toprule
\textbf{Model} & \textbf{Calibration} & \textbf{$\eta$} & \textbf{Split-half $K_A/K_B$} & \textbf{Rank $\rho_s$} & \textbf{Mask Jaccard} & \textbf{Full $K$} \\
\midrule
GaussianWorld lifecycle & 380 streaming frames & 0.8406 & 20,703 / 21,065$\dagger$ & 0.776$\dagger$ & 0.859$\dagger$ & 22,427 / 25,600 \\
GaussianFormer-1 & 500 frames & 0.6593 & 17,442 / 18,019 & 0.983 & 0.928 & 17,365 / 25,600 \\
GaussianFormer3D & 500 frames & 0.5688 & 24,478 / 24,506 & 0.975 & 0.960 & 22,535 / 25,600 \\
OccFormer & 200 frames & 0.7950 & 21 / 22 & 0.779 & 0.955 & 21 / 100 \\
GaussianFlowOcc & 300 scored / 8,971 forwarded & 0.8556 & 8,517 / 8,522 & 0.985 & 0.959 & 8,628 / 10,000 \\
\bottomrule
\end{tabular}%
}
\end{table}

$\dagger$ GaussianWorld split statistics are from an independent strict rerun over the same ten 38-frame training streams; its full $K=22{,}463$ differs from the primary $22{,}427$ by 36 rows.

All five score arrays in the automatic-selector study reproduce their reported budgets. Split-half mask Jaccard is 0.859--0.960. Rank agreement is lower for the streaming GaussianWorld rerun (0.776) and OccFormer (0.779) than for the other models (0.975--0.985), yet their masks remain stable at 0.859 and 0.955. GF3D illustrates a distinction: the two half-sample masks agree strongly with one another, yet their guarded-Otsu cuts are more conservative than the full-sample cut. We therefore report threshold, $K$, rank, and mask stability separately rather than treating one statistic as an identity gate. The provenance JSON/CSV records source identifiers.

The saved per-frame matrices also permit a prefix calibration-size sweep without rerunning or reading validation labels (Table~\ref{tab:calibration-sweep}).

\begin{table}[t]
\caption{Calibration-size sweep: survivor count and split-half stability versus scored frames}\label{tab:calibration-sweep}
\centering
\footnotesize
\setlength{\tabcolsep}{3pt}
\fitwidth{%
\begin{tabular}{lrrrrrr}
\toprule
\textbf{Model} & \textbf{Frames} & \textbf{$K$} & \textbf{$\eta$} & \textbf{Split-half $K_A/K_B$} & \textbf{$\rho_s$} & \textbf{Jaccard} \\
\midrule
GaussianFormer-1 & 100 & 18,383 & 0.601 & 22,234 / 20,697 & 0.925 & 0.877 \\
GaussianFormer-1 & 200 & 18,381 & 0.630 & 19,700 / 18,770 & 0.959 & 0.902 \\
GaussianFormer-1 & 500 & 17,365 & 0.659 & 17,442 / 18,019 & 0.983 & 0.928 \\
GaussianFormer3D & 100 & 23,365 & 0.794 & 22,226 / 21,980 & 0.909 & 0.892 \\
GaussianFormer3D & 200 & 24,076 & 0.735 & 23,103 / 23,239 & 0.947 & 0.925 \\
GaussianFormer3D & 500 & 22,535 & 0.569 & 24,478 / 24,506 & 0.975 & 0.960 \\
\bottomrule
\end{tabular}%
}
\end{table}

Ranking and mask agreement improve with more frames, but $K$ does not converge monotonically: relative to 500 frames, the 100/200-frame selections differ by 1,018/1,016 rows on GF1 and 830/1,541 rows on GF3D. This supports the ranking signal while showing that the automatically chosen operating point is calibration-size dependent.

Exact scoring is vectorized over candidate rows rather than implemented as 25,600 physical forwards; Table~\ref{tab:scoring-time} gives the wall times.

\begin{table}[t]
\caption{Scoring wall time per model}\label{tab:scoring-time}
\centering
\footnotesize
\setlength{\tabcolsep}{3pt}
\fitwidth{%
\begin{tabular}{lrr}
\toprule
\textbf{Model} & \textbf{Calibration work} & \textbf{Scoring wall time} \\
\midrule
GaussianWorld lifecycle & 10$\times$38-frame streams & 1.8 min \\
GaussianFormer-1 & 500 frames & 30.9 min \\
GaussianFormer3D & 500 frames & 31.5 min \\
OccFormer & 200 frames & 3.8 min (L4) \\
GaussianFlowOcc & 300 scored among 8,971 sequential forwards & 13.9 min \\
\bottomrule
\end{tabular}%
}
\end{table}

Times are single-run wall times from the saved A100 scorer logs, except OccFormer, which was scored on one L4. They include model forwarding and exact vectorized readout interventions, but exclude compilation, validation, and recovery. Cross-model timing is therefore descriptive rather than a hardware-normalized comparison.

\setcounter{table}{0}\renewcommand{\thetable}{B\arabic{table}}
\section{Matched recovery seed control}\label{sec:matched-recovery}

\subsection{GaussianFormer-1 three-seed matched recovery}\label{sec:matched-recovery-gf1}

The $K=17{,}365$ mask is frozen before recovery. Full and compiled arms use the same one-epoch recipe, and every evaluation covers all 753 validation batches.

\begin{table}[t]
\caption{GaussianFormer-1 three-seed matched recovery}\label{tab:gf1-three-seed}
\centering
\footnotesize
\setlength{\tabcolsep}{3pt}
\fitwidth{%
\begin{tabular}{lrrr}
\toprule
\textbf{Recovery seed} & \textbf{Full mIoU / geometry IoU} & \textbf{Compiled mIoU / geometry IoU} & \textbf{Compiled $-$ full} \\
\midrule
0 & 19.2927 / 30.0494 & 19.1198 / 29.8044 & -0.1730 / -0.2450 \\
1 & 19.2883 / 30.0527 & 19.1168 / 29.8051 & -0.1715 / -0.2476 \\
2 & 19.2907 / 30.0490 & 19.1225 / 29.8024 & -0.1681 / -0.2466 \\
\textbf{Mean $\pm$ sample std.} & \textbf{19.2906$\pm$0.0022 / 30.0503$\pm$0.0021} & \textbf{19.1197$\pm$0.0029 / 29.8040$\pm$0.0014} & \textbf{-0.1709$\pm$0.0025 / -0.2464$\pm$0.0013} \\
\bottomrule
\end{tabular}%
}
\end{table}

The mean semantic gap is 0.171 point and the largest observed gap across the three reported seeds is 0.173 point. The matched gap is much larger than its seed spread, so the small reduction is repeatable rather than an artifact of one recovery seed.

\setcounter{table}{0}\renewcommand{\thetable}{C\arabic{table}}
\section{GaussianWorld temporal selection diagnostics}\label{sec:temporal}

All results below are mechanism diagnostics at fixed $K=16{,}461$, evaluated on paired 400-frame subsets. They do not replace the 5,700-frame lifecycle-compiled result in Appendix~\ref{sec:gw-results}.

\subsection{Static, lifecycle-integrated, and composite scoring}

\begin{table}[t]
\caption{Static versus lifecycle-integrated scoring}\label{tab:static-vs-lifecycle}
\centering
\footnotesize
\setlength{\tabcolsep}{3pt}
\fitwidth{%
\begin{tabular}{lrrr}
\toprule
\textbf{Fixed-$K$ ranking} & \textbf{Semantic mIoU} & \textbf{Geometry IoU} & \textbf{Occupied-voxel argmax drift} \\
\midrule
Static Mean-FLIP & 20.041 & 33.626 & 19.60\% \\
Lifecycle Mean-FLIP & 20.304 & 34.064 & 20.27\% \\
Lifecycle composite (mean/CVaR/frequency) & 20.343 & 34.086 & 20.15\% \\
\bottomrule
\end{tabular}%
}
\end{table}

Lifecycle-aware rankings improve subset mIoU and geometry IoU at the same budget, although occupied-voxel drift does not improve monotonically. This is why the paper uses the direct lifecycle integral and guarded Otsu rather than promoting the exploratory composite.

\subsection{Fixed compilation and causal routing}

\begin{table}[t]
\caption{Fixed masks and dynamic routers}\label{tab:routers}
\centering
\footnotesize
\setlength{\tabcolsep}{3pt}
\fitwidth{%
\begin{tabular}{lrrrrr}
\toprule
\textbf{Readout policy at $K=16{,}461$} & \textbf{$\Delta$ mIoU vs. paired full} & \textbf{$\Delta$ Geometry IoU} & \textbf{Occupied drift} & \textbf{Mean-FLIP mass recall} & \textbf{Latency reduction} \\
\midrule
Static fixed mask under full history & -10.823 & -22.090 & 77.15\% & --- & --- \\
Per-frame Decision-FLIP oracle, full history & -0.157 & -0.489 & 1.82\% & 100\% & oracle cost excluded \\
Vote-mass router, free rollout & -0.417 & -1.358 & 18.27\% & --- & 0.51\% \\
Learned MLP router, free rollout & -0.372 & -0.492 & 14.76\% & 98.43\% & 1.82\% \\
Causal selective-SSM router, free rollout & -0.804 & -1.092 & 18.73\% & 96.86\% & 0.07\% \\
\bottomrule
\end{tabular}%
}
\end{table}

The static mask's large drift shows that a row selected on isolated readout evidence need not retain the same role after recurrent state evolution. Dynamic routers recover much of the oracle ranking signal, but keeping all 25,600 rows in temporal state and routing only before aggregation yields little end-to-end speedup. These diagnostics motivate lineage-aware \emph{physical compilation} rather than a runtime router.

\end{document}